\documentclass[conference]{IEEEtran}
\IEEEoverridecommandlockouts
\usepackage{cite}
\usepackage{amsmath,amssymb,amsfonts}
\usepackage{algorithmic}
\usepackage{graphicx}
\usepackage{textcomp}
\usepackage{xcolor}
\usepackage{svg}
\graphicspath{ {./images/} }
\usepackage[numbers]{natbib}
\usepackage{float}
\usepackage{titlesec}
\usepackage{ulem}
\usepackage{tabularx}
\usepackage{makecell}
\usepackage[T1]{fontenc}
\usepackage{multirow}

\usepackage{soul} 
\usepackage{subcaption}

\usepackage{enumitem}
\setlist{leftmargin=5.5mm}

\DeclareUnicodeCharacter{2212}{-}

\makeatletter
\newcommand{\mathleft}{\@fleqntrue\@mathmargin0pt}
\newcommand{\mathcenter}{\@fleqnfalse}
\makeatother

\newcommand{\vspacecontrol}[0]{\vspace{-.3cm}}

\DeclareMathOperator*{\argmin}{arg\,min}

\begin{document}

\title{
 Deep Reinforcement Learning Versus Evolution Strategies: A Comparative Survey
}

\makeatletter
\newcommand{\linebreakand}{%
  \end{@IEEEauthorhalign}
  \hfill\mbox{}\par
  \mbox{}\hfill\begin{@IEEEauthorhalign}
}
\makeatother

\author{
\IEEEauthorblockN{Amjad Yousef Majid}
\IEEEauthorblockA{\textit{Delft University of Technology} \\
a.y.majid@tudelft.nl}
\and
\IEEEauthorblockN{Serge Saaybi}
\IEEEauthorblockA{\textit{Delft University of Technology} \\
s.c.e.saaybi@student.tudelft.nl}
\and
\IEEEauthorblockN{Tomas van Rietbergen}
\IEEEauthorblockA{\textit{Delft University of Technology} \\
T.L.vanRietbergen@student.tudelft.nl}
\linebreakand
\IEEEauthorblockN{Vincent Francois-Lavet}
\IEEEauthorblockA{\textit{VU Amsterdam} \\
vincent.francoislavet@vu.nl}
\and
\IEEEauthorblockN{R Venkatesha Prasad}
\IEEEauthorblockA{\textit{Delft University of Technology} \\
r.r.venkateshaprasad@tudelft.nl}
\and
\IEEEauthorblockN{Chris Verhoeven}
\IEEEauthorblockA{\textit{Delft University of Technology} \\
C.J.M.Verhoeven@tudelft.nl} \hfill}

\maketitle
\begin{abstract}
    Deep Reinforcement Learning (DRL) and Evolution Strategies (ESs) have surpassed human-level control in many sequential decision-making problems, yet many open challenges still exist.
To get insights into the strengths and weaknesses of DRL versus ESs, an analysis of their respective capabilities and limitations is provided. 
After presenting their fundamental concepts and algorithms, a comparison is provided on key aspects such as scalability, exploration, adaptation to dynamic environments, and multi-agent learning. 
Then, the benefits of hybrid algorithms that combine concepts from DRL and ESs are highlighted. 
Finally, to have an indication about how they compare in real-world applications, a survey of the literature for the set of applications they support is provided.

\end{abstract}


\section{Introduction}
In the biological world, the intellectual capabilities of humans and animals have developed through a combination of evolution and learning. 
On the one hand, evolution has allowed living beings to improve genetically over successive generations such that higher forms of intelligence have appeared, on the other hand, adapting rapidly to new situations is possible due to the learning capability of animals and humans.

In the race for developing artificial general intelligence, these two phenomena have motivated the development of two distinct approaches that could both play an important role in the quest for intelligent machines.
From the learning perspective, \textit{Reinforcement learning (RL)} shows many parallels with how humans and animals can deal with new unknown sequential decision-making tasks.
Meanwhile, \textit{Evolution Strategies (ESs)} are engineering methods inspired by how the mechanism that let intelligence emerge in the biological world---repeatedly selecting the best performing individuals.

In this paper, we discuss RL and ESs together analyzing their strengths and weaknesses regarding their sequential decision-making capabilities and shed light on potential  directions for further development. 

The RL framework is formalized as an agent acting on an environment with the goal of maximizing a cumulative reward over the trajectory of interaction with the environment \cite{sutton2018reinforcement}. Imagine playing a table tennis game (environment) with a robot (agent). The robot has not explicitly been programmed to play the game. Instead, it can observe the score of the game (rewards). The robot's goal is to maximize its score. For that purpose, it tries different techniques of hitting the ball (actions), observes the outcome, and gradually enhances its playing strategy (policy). 

Despite the proven convergence of RL algorithms to optimal policies---best solutions to the problems at hand---they face difficulties processing high-dimensional data (e.g., images). 
To tackle problems with high-dimensional data, RL algorithms are nowadays often combined with deep neural networks, giving raise to a whole field of research known as Deep RL (DRL)~\cite{Vincent2018}.

\begin{figure}[t]
    \centering 
    \includegraphics[width=\columnwidth]{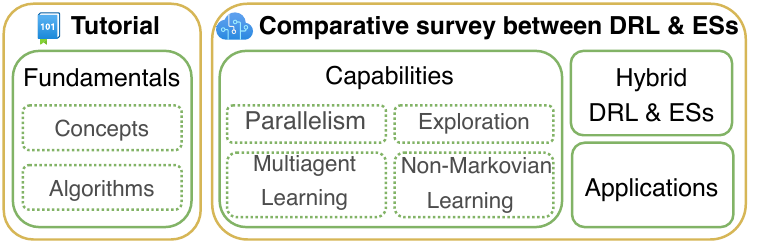}
    \caption{The structure of the survey}
    \label{fig_intro}
    \vspacecontrol
\end{figure}

As a contrasting approach to DRL, ES algorithms utilize a random process to iteratively generate candidate solutions. Then, they evaluate these solutions and bias the search in direction of the best scoring ones \cite{EibenS03}. In recent years, ESs have seen an increase in popularity and has been successfully applied to several applications, including optimizing objective functions for many RL tasks \cite{salimans2017evolution, chrabaszcz2018basics}. 
\begin{figure*}[!t]
    \begin{subfigure}[b]{0.33\textwidth}
        \centering
    \includegraphics[width=\columnwidth]{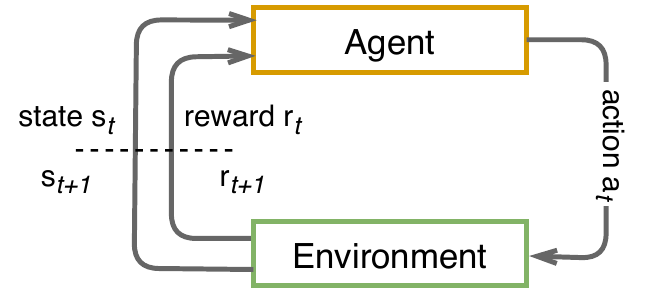}
    \caption{RL}
    \label{fig:drl_problem_setup}
    \end{subfigure}
    \begin{subfigure}[b]{0.33\textwidth}
        \centering
        \includegraphics[width=\columnwidth]{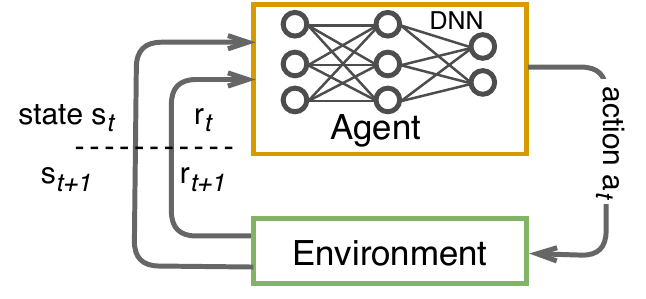}
        \caption{DRL}
        \label{fig:DRL_fig}
    \end{subfigure}
    \begin{subfigure}[b]{0.33\textwidth}
        \centering 
        \includegraphics[width=\columnwidth]{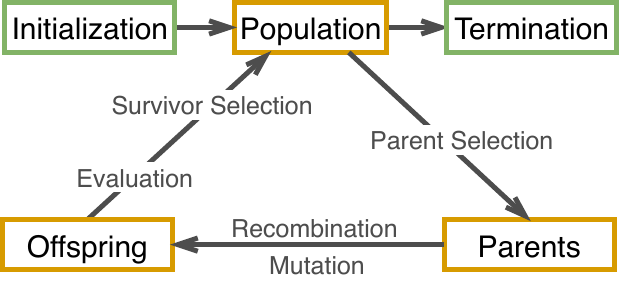}
        \caption{ESs}
        \label{fig:ESs}
    \end{subfigure}
    \caption{Iteration loops of (Deep) Reinforcement Learning and Evolutionary Strategies.}
    \label{fig:qlearningVSsarsa}
    \vspacecontrol
\end{figure*}

The parallel development of DRL and ESs indicates that each has its advantages (and disadvantages), depending on the problem setup. To enable scientists and researchers to choose the best algorithm for the problem at hand, 
we summarized the pros and cons of these approaches through the development of a comparative survey: we compared DRL and ESs from different learning aspects such as scalability, exploration, the ability to learn in dynamic environments and from an application standpoint (Figure~\ref{fig_intro}). We also discuss how combining DRL and ESs in hybrid systems can leverage the advantages of both approaches. 

To date, there have been different papers summarizing different features of DRL and ESs.
For example, derivative-free reinforcement learning (e.g., ESs) has been reviewed in \cite{Qian2021}, covering aspects such as scalability and exploration. A survey related to DRL for autonomous driving is provided in \cite{Kiran2021}, and  the challenges, solutions, and applications of multi-agent DRL systems are reviewed in \cite{nguyen2020deep}. However, contrasting with prior work, our paper surveys the literature with a bird's-eye view, focusing on the main developmental directions instead of individual algorithms. 

The rest of the paper is organized  as  follows:  Section~\ref{sec:fund_comp} presents the fundamental architectural concepts behind RL and ESs; Section~\ref{sec:fund_alg} summarizes fundamental algorithms of RL, DRL and ESs; Sections~\ref{sec:parallel}, \ref{sec:exploration}, \ref{sec:dynamic} and  \ref{sec:multiagent} compare the capabilities of DRL and ESs; In Section~\ref{sec:hybrid}, we present hybrid systems that combine DRL and ESs. Section ~\ref{sec:applications} compares them from an applications' point of view. Section ~\ref{sec:future_direction} outlines open challenges and potential research directions. Finally, we conclude the paper in Section~\ref{sec:conclusion}. The main takeaways of each section are summarized in a concise sub-section titled ``Comparison''.

\section{Fundamentals}
\label{sec:fund_comp}
This section covers the fundamental elements of DRL and ESs, including formal definitions and the main algorithmic families. 
\subsection{Reinforcement Learning}
Reinforcement Learning (RL) is a computational approach to understanding and automating goal-directed 
learning and decision making~\cite{sutton2018reinforcement}. The goal of an RL agent is to 
maximize the total reward it receives when interacting with an environment 
(Figure~\ref{fig:drl_problem_setup}), which is generally modeled as a 
Markov Decision Process (MDP). An MDP is defined by the tuple 
$(\mathcal{S},\mathcal{A},T, R)$, where $\mathcal{S}$ denotes the state space; $\mathcal{A}$ is the action space; $T(s,a,s')$ is a transition function that defines the 
probability of transitioning from the current state $s$ to the next state 
$s'$ after an agent takes action $a$; $R(s,a,s')$ is the reward function that 
defines the immediate reward $r$ that the agent observes after taking 
action $a$ and the environment transition from $s$ to $s'$.

The total \textit{return} starting from time $t$ until the end of the interaction 
between an agent and its environment is expressed as
$$ 
G_t = \sum_{k=0}^{\infty} \gamma^k R_{t+k+1} \text{,} 
$$
where $R_ t$ and is a random variable that models the immediate reward, $r$, and $\gamma \in [0,1)$ is a discount factor that weights the immediate and future rewards. \textit{Value functions} are the expected return of being in a state or taking a particular action. The state-value function $v_{\pi}(s)$ gives the expected return from state $s$ following policy $\pi$, 
\begin{equation}
    v^{\pi}(s) = \sum_{a}\pi(a|s) \sum_{s',r} p(s',r|s,a)[r + \gamma v^{\pi}(s')] \text{.}
    \label{eq:state-value-function}
\end{equation}
The action-value function (or Q-function) $q^{\pi}(s,a)$ is the expected return of taking action $a$ in state $s$ and following policy $\pi$ thereafter, 
\begin{equation}
q^{\pi}(s,a) =  \sum_{s',r} p(s',r|s,a)\bigg[r + \gamma \sum_{a'}\pi(a'|s') q^{\pi}(s',a')\bigg] \text{.}
    \label{eq:action-value-function}
\end{equation}
The action selection process of an agent is governed by its policy, which in the general stochastic case yields an action according to a probability distribution over the action space conditioned on a given state $\pi(s,a)$.

\noindent There are four main RL algorithmic families: 

\textit{Policy-based Algorithms}.
A policy-based algorithm optimizes and memorizes a policy explicitly, that is, it directly searches the policy space for an (approximate) optimal policy, $\pi^*$. Examples of such algorithms are policy iteration~\cite{policyiter}, policy gradient~\cite{policygradient} and REINFORCE~\cite{WilliamsREINFORCE1992}. Policy-based algorithms can be applied to any type of action space: continuous, discrete or a mixture (multiactions). However, these algorithms generally have high variance and are sample-inefficient.

\textit{Value-based Algorithms}. 
A value-based algorithm learns a value function, $ v^{\pi}(s) \text{ or } q^{\pi}(s,a)$. Then, a policy is extracted according to the learned value function. Examples of such algorithms are value iteration~\cite{valueiter}, SARSA~\cite{SARSA}, Q-learning and DQN~\cite{mnih2013playing}. Value-based algorithms are more sample-efficient than policy-based ones. However, under ordinary circumstances the convergence of these algorithms is not guaranteed.

\textit{Actor-critic-based Algorithms}.
The actor-critic approach tries to combine the strengths of policy- and value-based algorithms into a single algorithmic architecture~\cite{konda2000actor}. The actor is a policy-based algorithm that tries to learn the optimal policy, whereas the critic is a value-based algorithm that evaluates the actions taken by the actor. 

\textit{Model-based Algorithms}. All of the algorithmic families mentioned previously concern 
\textit{model-free} algorithms. In contrast, model-based algorithms learn or make use of a model of the transition dynamics of an environment. Once an agent has access to such a model, it can use it to 
``imagine'' the consequences of taking a particular set of actions without acting on the environment. Such capability enables an RL agent to evaluate the expected actions of an opponent in games~\cite{kaiser2019model, Silver2016Go} and to make better use of gathered data, which is very useful in tasks such as controlling a robot~\cite{polydoros2017survey}. However, for many problems,
it is difficult to produce close to reality models. 

\textbf{Deep Reinforcement Learning} (DRL) refers to the combination of Deep Learning (DL) and RL (Figure~\ref{fig:DRL_fig}) \cite{Vincent2018}. DRL uses DNNs to approximate one of the learnable functions of RL. Correspondingly, there are three main families of DRL algorithms: value-based, policy-based, and model-based~\cite{mnih2013playing,lillicrap2019continuous,kaiser2019model}. 
For example, the DNN of a policy-based DRL agent takes the
state of the environment as input and produces an action as output (Figure \ref{fig:DRL_fig}). The action selection process is governed by the parameters $\pmb{\theta}$ of the DNN. 
The parameters selection is optimized using a backpropagation algorithm during the training phase. 

\subsection{Evolution Strategies} 
Evolution Strategies (ESs) are set of a population-based black-box optimization algorithms often applied to continuous search spaces problems to find the optimal solutions \cite{hansen2015,Zhenhua2020}.  
ESs do not require modeling the problem as an MDP, neither the objective function $f(\pmb{x})$ has to be differentiable and continuous. The latter explains why ESs are gradient-free optimization techniques. They do however require the objective function $f(\pmb{x})$ to be able to assign a fitness value to (i.e., to evaluate) each input $\pmb{x} \in \mathbb{R}^n$ such that $f : \mathbb{R}^n \rightarrow \mathbb{R}$, $\pmb{x} \rightarrow f(\pmb{x})$.

The basic idea behind ESs is to bias the sampling process of candidate solutions towards the best individuals found so far until a satisfactory solution is found. 
Samples can be drawn for instance from a (multivariate) normal distribution whose shape (i.e., the mean $m$ and the standard deviation $\sigma$) is described by what are called \textit{strategic parameters}. These can be modified online to make the search process more efficient. The generic ESs process is shown in Figure~\ref{fig:ESs} and its elements are explained below:
\begin{enumerate}
    \item \textit{Initialization}: the algorithm generates an initial population $P$ consisting of $\mu$ individuals.
    \item \textit{Parent selection}: a sub-set of the population is selected to function as parents during the recombination step.
    \item \textit{Reproduction} consists of two steps:
        \begin{enumerate}
            \item \textit{Recombination}: two or more parents are combined to produce a mean for the new generation.
            \item \textit{Mutation}: a small amount of noise is added to the recombination results. A common way of implementing mutation is to sample from a multivariate normal distribution centered around the mean obtained from the previous recombination step:
            $$\pmb{x}_k^{g+1} \sim \mathcal{N}(\pmb{m}^{(g)},\sigma^{(g)}I)=\pmb{m}^{(g)}+\sigma^{(g)}\mathcal{N}(0,I)\text{,}$$ 
            where $g$ is the generation index,  $k$ is the number of offsprings, and $I$ is the identity matrix.
        \end{enumerate}
    \item \textit{Evaluation}: a fitness value is assigned to each candidate solution using the objective function $f(x_i)$.
    \item  \textit{Survivor selection}: the best $\mu$ individuals are selected to form the population for the next generation. Generally, the algorithm iterates from step 2 to step 5 until a satisfactory solution is found.
\end{enumerate}
The idea of employing ESs as an alternative to RL is not new \cite{Verena2008,Meisner2008,HEIDRICHMEISNER2009152,Meisner2009}, but recently it has seen a renewed interest (e.g.  \cite{salimans2017evolution, Chrabaszcz2018}). 

\subsection{Comparison}
Our main takeaways of the above fundamental concepts are: 
\begin{itemize}
    \item The objective of an RL algorithm is to maximize the sum of discounted rewards, whereas an ESs algorithm does not require such formulation. However, the objective for RL settings can be converted to ESs settings with a terminal state that provides a reward equivalent to the fitness function.
    \item The problem setup differs between RL and ESs. An ESs algorithm is a black-box optimization method that keeps a pool of multiple candidate solutions, while an RL method generally has a single agent that improves its policy by interacting with its environment. 
    \item An ESs algorithm aims at finding candidate solutions that optimize a fitness function, whereas the goal of DRL is to keep advancing one or two function approximators which in turn need to optimize the equivalent of the fitness function, usually defined by the discounted return.
    \item The ESs approach is most similar to the policy-based DRL approach: both aim at finding parameters in a search space such that the resulting parameterized function optimizes certain objectives (expected return for DRL or fitness score for ESs). The main distinction is that ESs, unlike DRL, do not calculate gradients nor use backpropagation. 
    \item Value-based RL methods usually operate in discrete action spaces while the actor-critic architecture extends this ability to continuous action spaces. ESs can operate on discrete or continuous action spaces by default.
\end{itemize}

\section{Fundamental Algorithms}
\label{sec:fund_alg}
\begin{table*}[t]
\centering
    \caption{Fundamental (Deep) Reinforcement Learning and Evolution Strategies algorithms}
    \begin{tabularx}{\textwidth}{ p{1.4cm}|p{2cm}|p{2.2cm}|p{3.6cm}|p{4.4cm}|X|p{.3cm}}
         \hline  \textbf{Algorithm} & \textbf{Classification} & \textbf{Action Space} & \textbf{Memory Consumed} & \textbf{Limitations} & \textbf{Backprop.} & \textbf{Ref.} \\
            \hline
            SARSA & on-policy value-based RL & discrete & exponential in state and action spaces & tackling continuous space, does not generalize between similar states & \text{\sffamily X} & \cite{sutton2018reinforcement}\\ 
            \hline
            Q-learning & off-policy value-based RL & discrete & exponential in state and action spaces & tackling continuous space, does not generalize between similar states  & \text{\sffamily X} & \cite{Watkins1992}\\
            \hline
            REINFORCE & policy-based RL & discrete/continuous & typically, it requires storing DNN parameters & 
            data inefficiency, higher variance compared to DQN & \checkmark & \cite{WilliamsREINFORCE1992} \\
            \hline
            DQN &  off-policy value-based RL & discrete &
            it requires storing DNN parameters and a replay buffer
            & the learning of the Q-function can suffer from instabilities & \checkmark & \cite{Mnih2015} \cite{Vincent2018}\\
            \hline
            CMA-ES & black-box ES optimization & discrete/continuous & high memory requirement 
            & high space and time complexity when dealing with large scale optimization problems  & \text{\sffamily X} & \cite{ijcai2019} \cite{Varelas2018} \\
            \hline
            NES \& OpenAI-ES & black-box ES optimization & discrete/continuous & less memory usage than CMA-ES
            & data inefficiency due to gradient approximation &  \text{\sffamily X} &  \cite{salimans2016improved} \cite{conti2018improving}\\
            \hline
    \end{tabularx}
    \label{tab:algo_comp}
\end{table*}
Fundamental algorithms of (D)RL and ESs are introduced in this section. 

\subsection{Reinforcement Learning Algorithms}
\noindent\textbf{SARSA} is a model-free algorithm that leverages temporal-differences for prediction~\cite{sutton2018reinforcement}. It updates the Q-value, $Q(s_t,a_t)$,  while following a policy. The interaction between the agent and environment results in the following sequence $\dots,s_t,a_t,r_{t+1},s_{t+1},a_{t+1},\dots$: the agent takes an action $a_t$ while being in a state $s_t$, and consequently, the environment transitions to a state $s_{t+1}$ and the agent observes a reward $r_{t+1}$.  For action selection, SARSA uses $\varepsilon$-greedy algorithm, which
selects the action with maximum $Q(s_t,a_t)$ with probability of $1-\varepsilon$, and otherwise, it draws an action uniformly from $\mathcal{A}$. SARSA is an on-policy algorithm, that is, it evaluates and improves the same policy that selects the taken actions. SARSA's update equation is
\begin{multline}
    Q(s_t,a_t)\leftarrow Q(s_t,a_t)+\alpha \big[r_{t+1}+ \\  \gamma Q(s_{t+1},a_{t+1})-Q(s_t,a_t) \big] \text{,}
    \label{eq_sarsa}
\end{multline}
where $\alpha$ is the learning rate.

\noindent\textbf{Q-Learning}~\cite{sutton2018reinforcement} is similar to SARSA with a key difference: It is an off-policy algorithm, which means that it learns an optimal Q-value function from data obtained via any policy (without introducing a bias). 
In particular, The Q-learning update rule compares the Q-value of the current state-action pair, $Q(s_t,a_t)$, with a pair from the next state that has the maximum Q-value, $Q(s_{t+1},a_{t+1})$, which is not necessarily the one chosen by  $\varepsilon$-greedy as in SARSA.    
The update rule of Q-learning is 
\begin{multline}
    Q(s_t,a_t)\leftarrow Q(s_t,a_t)+\alpha \big[r_{(t+1)}+\\
    \gamma \max_a Q(s_{(t+1)},a_{(t+1)})-Q(s_t,a_t) \big ] \text{.}
    \label{eq_qlearning}
\end{multline}

Off-policy algorithms are more data-efficient than on-policy ones, because they can use the collected data repeatedly. 

\noindent\textbf{REINFORCE}~\cite{WilliamsREINFORCE1992} is a fundamental stochastic gradient descent algorithm for policy gradient algorithms. It leverages a DNN to approximate the policy $\pi$ and update its parameters $\pmb{\theta}$. 
The network receives an input from the environment and outputs a probability distribution over the action space, $\mathcal{A} $. The steps involved in the implementation of REINFORCE are:
\begin{enumerate}
    \item Initialize a Random Policy (i.e., the parameters of a DNN)
    \item Use the policy $\pi_{\pmb{\theta}}$ to collect a trajectory  $\tau=(s_0,a_0,r_1,s_1,a_1,r_2,...,a_H,r_{H+1},s_{H+1})$
    \item Estimate the return for this trajectory
    \item Use the estimate of the return to calculate the policy gradient:
    \begin{equation}
    \nabla_{\pmb{\theta}} \mathcal{J}(\pmb{\theta}) = \mathbb{E}_{\pi_{\pmb{\theta}}} [\nabla \log \pi(a \vert s; \pmb{\theta}) Q_\pi(s, a)]
    \label{eq_grad}
    \end{equation}
    \item Adjust the weights $\pmb{\theta}$ of the Policy: \\
    $\pmb{\theta} \leftarrow \pmb{\theta} + \alpha\nabla_\theta J(\pmb{\theta})$
    \item Repeat from step 2 until  termination.
\end{enumerate}
 
\noindent\textbf{Deep Q-network (DQN)}~\cite{Mnih2015} combines Q-learning with a convolutional neural network (CNN)~\cite{krizhevsky2012imagenet} to act in environments with high-dimensional input spaces (e.g., images of Atari games). It gets a state (e.g., a mini-batch of images) as input and produces Q-values of all possible actions. The CNN is used to approximate the optimal action-value function (or Q-function). Such usage, however, causes the DRL agent to be unstable~\cite{tsitsiklis1997analysis}. To counter that, DQN samples an experience replay~\cite{lin1993reinforcement} dataset $D_t=\{(s_1,a_1,r_2,s_2), \dots, (s_t,a_t,r_{t+1},s_{t+1})\}$ and uses a target network that is updated only after a certain number of iterations. To update the network parameters at iteration $i$, DQN uses the following loss function 
\begin{multline}
    L_i(\theta_i)=\mathbb{E}_{(s,a,r,s')} \sim U(D) \\
    \bigg[\left(r+\gamma \max_{a'} Q(s',a';\overline{\pmb{\theta}_i}) - Q(s',a';\pmb{\theta}_i) \right)^2 \bigg]
    \label{eq:dqn}
\end{multline}
where $\theta_i$ and $\overline{\theta_i}$ are the parameters of the Q-network and target network, respectively; and the experiences, $(s,a,r,s')$, are drawn from $D$ uniformly. 

\subsection{Evolutionary Strategies Algorithms}

\noindent The (1+1)\textbf{-ES} (one parent, one offspring) is the simplest ES conceived by~\citet{Rechenberg_1973}. 
First, a parent candidate solution, $\textbf{x}_p$, is drawn according to a uniform random distribution from an initial set of solutions, $\{\textbf{x}_i, \textbf{x}_j \}$. The selected parent, $\textbf{x}_p$, together with its fitness values enter the evolution loop. In each generation (or iteration) an offspring candidate solution, $\textbf{x}_o$, is created by adding a vector drawn from an uncorrelated multivariate normal distribution to $\textbf{x}_p$ as follows:  
$$ 
\textbf{x}_o = \textbf{x}_p + \textbf{y}\sigma, \textbf{y} \sim \mathcal{N}(0,\textbf{I})\text{.} 
$$ 
If the offspring $\textbf{x}_o$ is found to be fitter than the parent $\textbf{x}_p$ then it becomes the new parent for the next generation, otherwise it is discarded. This process is repeated until a termination condition is met. The amount of mutation (or perturbation) added to $\textbf{x}_p$ is controlled by the stepsize parameter $\sigma$. The value of $\sigma$ is updated every predefined number of iterations according to the well-known $\frac{1}{5th}$ success rule
~\cite{Slowik2020, Dianati2002AnIT}: 
if $\textbf{x}_o$ is fitter than $\textbf{x}_p$ $\frac{1}{5th}$ of the times then $\sigma$ should stay the same; if $\textbf{x}_o$ is fitter \textit{more} than $\frac{1}{5th}$ of the times then $\sigma$ should be increased, and otherwise it should be decreased. 

\noindent The ($\mu/\rho\overset{+}{,}\lambda$)\textbf{-ES} was originally proposed by~\citet{sulfur1977numeric} as an extension to the (1+1)-ES. Instead of using one parent to generate one offspring, it uses $\mu$ parents to generate $\lambda$ offsprings using both recombination and mutation. 
In the comma-variation of this algorithm (i.e., ($\mu/\rho{,}\lambda$)-ES) the selection of the parents for the next generation happens solely from the offsprings. Whereas in the plus-variation, the selection of the parents for the next generation happens from the union of the offsprings and old parents. The $\rho$ in the name of the algorithm refers to the number of parents used to generate each offspring. 

An element (or an individual) that the ($\mu/\rho\overset{+}{,}\lambda$)-ES evolves consists of ($\textbf{x}$, $\textbf{s}$, $f$) where $\textbf{x}$ is the candidate solution, $\textbf{s}$ are the strategy parameters that control the significance of the mutation, and $f$ holds the fitness value of  $\textbf{x}$. Consequently, the evolution process itself tunes the strategy parameters which is known as self-adaptation. Thus, unlike (1+1)-ES, ($\mu/\rho\overset{+}{,}\lambda$) do not need external control settings to adjust the strategy parameters. 

\noindent\textbf{Covariance Matrix Adaptation Evolution Strategies (CMA-ES)} is one of the most popular gradient-free optimisation algorithms~\cite{Hansen1996,Hansen2001,Hansen2003,hansen2016cma}.
To search a solution space, it samples a population, $\lambda$,  of new search points (offsprings) from a multivariate normal distribution: 
$$\pmb{x}_i^{g+1} = \pmb{m}^{(g)}+\sigma^{(g)}\mathcal{N}(\pmb{0},\pmb{C}^{(g)}) \text{ for } i=1\text{, \dots ,}\lambda\text{,}$$ 
where $g$ is the generation number (i.e., $g=1,2,3,\dots$), $\pmb{x}_i \in \mathbb{R}^n$ is the $i$-th offspring, $\pmb{m}$ and $\sigma$ denote the mean and standard deviation of $\pmb{x}$, $\pmb{C}$ represents the covariance matrix, and $\mathcal{N}(\pmb{0},\pmb{C})$ is a multivariate normal distribution.  To compute the mean for the next generation, $\pmb{m}^{g+1}$, CMA-ES computes a weighted average of the best---according to their fitness values---$\mu$ candidate solutions, where $\mu < \lambda$ represents the parent population size. Through this selection and the assigned weights, CMA-ES biases the computed mean towards the best candidate solutions of the current population. 
It automatically adapts the stepsize $\sigma$ (the mutation strength) using the Cumulative Stepsize Adaption (CSA) algorithm~\cite{Hansen1996} and an evolution path, $\pmb{p}_{\sigma}$: if $\pmb{p}_{\sigma}$ is longer than the expected length of the evolution path under random selection $\mathbb{E} ||\mathcal{N}(\pmb{0}, \pmb{I})||$, then increase the stepsize; otherwise, decrease it. 
To direct the search towards promising directions, 
CMA-ES updates the covariance matrix in each iteration. The update consists of two main parts: (i) rank-1 update, which computes an evolution path for the mutation distribution means, similarly to the stepsize evolution path; and (ii) rank-$\mu$ update, which computes a covariance matrix as a weighted sum of covariances of the best $\mu$ individuals. The obtained results from these steps are used to update the covariance matrix $\pmb{C}$ itself. The algorithm iterates until a satisfactory solution is found (we refer the interested reader to ~\cite{hansen2016cma} for a more detailed explanation).

\noindent\textbf{Natural Evolution Strategies (NES)} is similar in many ways to the previously defined ES algorithms, the core idea behind it relates to the use of gradients to adequately update a search distribution \cite{wierstra2014natural}. The basic idea behind NES consists of: 
\begin{itemize}
  \item \textit{Sampling:} NES samples its individuals from a probability distribution (usually a Gaussian distribution) over the search space. The end goal of NES is to update the distribution parameters $\pmb{\theta}$ to maximize the average fitness $F(\pmb{x})$ of the sampled individuals $\pmb{x}$.
  \item \textit{Search gradient estimation:} NES estimates a search gradient on the parameters by evaluating the samples previously computed. It then decides on the best direction to take to achieve a higher expected fitness.
  \item \textit{Gradient ascent:} NES computes gradient ascent along the estimated gradient
  \item Iterates over the previous steps until a stopping criterion is met \cite{wierstra2014natural}.
\end{itemize}

\citet{salimans2017evolution} proposed a variant of NES for optimizing the policy parameters $\theta$. As gradients are unavailable, they are estimated via gaussian smoothing of the objective function $F(X)$ which represents the expected return. 

\subsection{Comparison}
Our main observations of the fundamental algorithms are:
\begin{itemize}
    \item Both ES and on-policy RL algorithms are data inefficient: on-policy algorithms make use of data that is generated from the current policy and discard older data; ES discard all but a sub-set of candidate solutions in each iteration.  
    \item The computation requirements per iteration of ESs are often lower than that of DRL as it does not require backpropagating error values.
    \item Value-based DRL algorithms such as DQN can be data-efficient because they can work with a replay memory that allows a reuse of off-policy data. However, they can become unstable for long horizons and high discount factors \cite{franccois2015discount}.
    \item Policy-based RL and ESs are similar in that they both search for good policies directly. 
    \item Table~\ref{tab:algo_comp} highlights important characteristics of the mentioned algorithms.
\end{itemize}

\section{Deep Reinforcement Learning Versus Evolution Strategies} 
This section compares different aspects of DRL and ESs, such as their 
ability to parallelize computations, explore an environment, and learn in multi-agent and dynamic settings. 
    \subsection{Parallelism}
    \label{sec:parallel}
\begin{figure*}[t]
    \begin{subfigure}[b]{0.32\textwidth}
        \centering
    \includegraphics[width=\columnwidth]{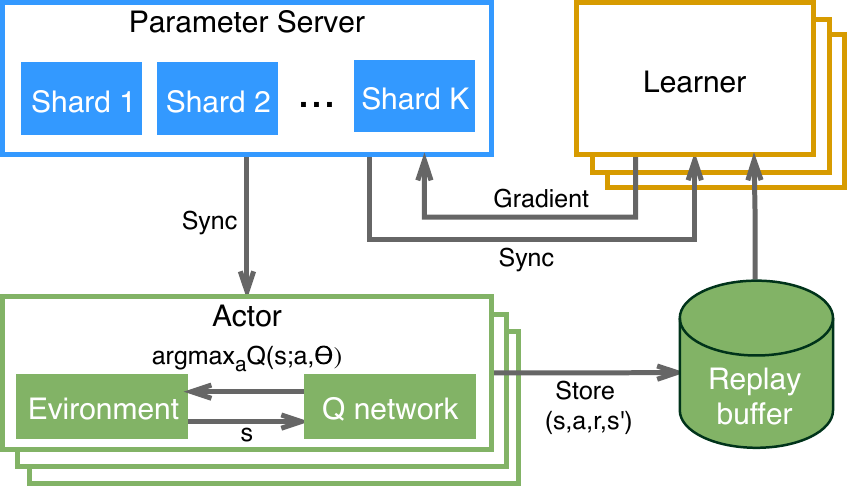}
    \caption{Gorila~\cite{nair2015massively} 
    }
    \label{fig_Gorila}
    \end{subfigure}
    \hfill
    \begin{subfigure}[b]{0.32\textwidth}
        \centering
        \includegraphics[width=\columnwidth]{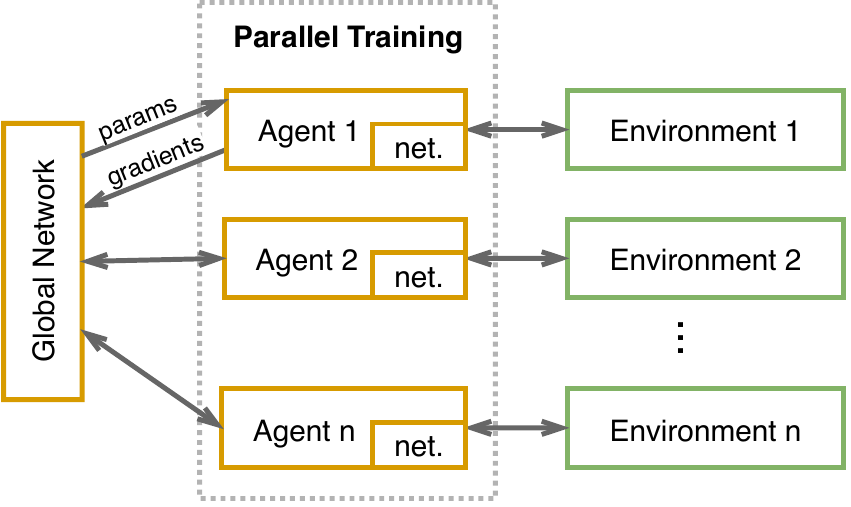}
        \caption{A3C~\cite{mnih2016asynchronous} 
        }
        \label{fig_A3C}
    \end{subfigure}
    \hfill
    \begin{subfigure}[b]{0.32\textwidth}
        \centering
        \includegraphics[width=\columnwidth]{images/batched_a2C.pdf}
        \caption{Batched A2C~\cite{clemente2017efficient} 
        }
        \label{fig_GPUpar}
    \end{subfigure}
    \par\bigskip
    \begin{subfigure}[b]{0.32\textwidth}
        \centering
    \includegraphics[width=\columnwidth]{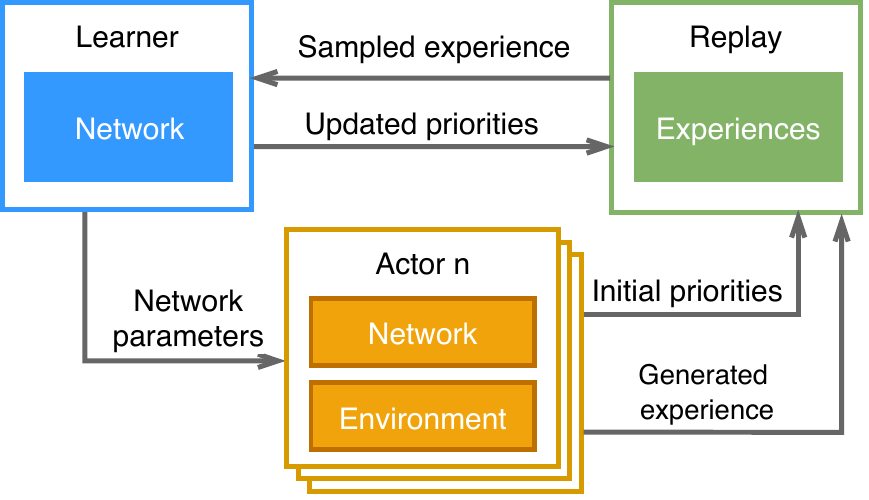}
    \caption{Ape-X~\cite{horgan2018distributed}
    }
    \label{fig_apex}
    \end{subfigure}
    \hfill
    \begin{subfigure}[b]{0.32\textwidth}
        \centering
        \includegraphics[width=\columnwidth]{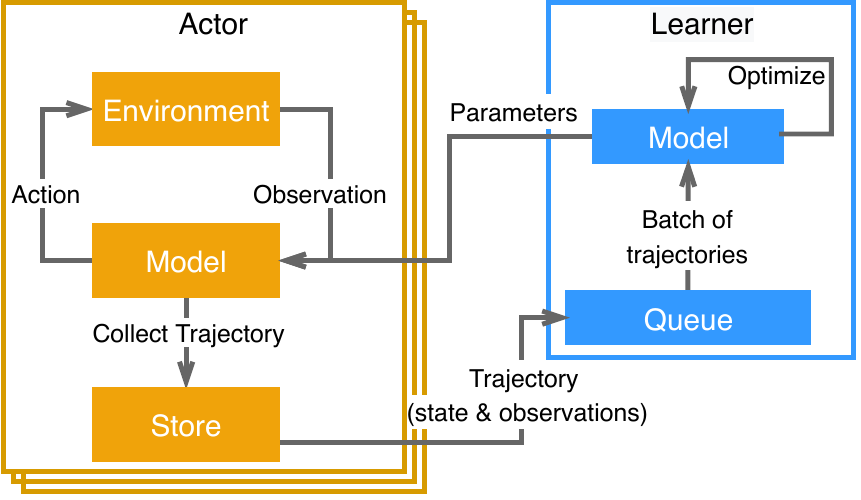}
        \caption{Impala~\cite{espeholt2018impala}
        } 

        \label{fig_impala}
    \end{subfigure}
    \hfill
    \begin{subfigure}[b]{0.32\textwidth}
        \centering
        \includegraphics[width=\columnwidth]{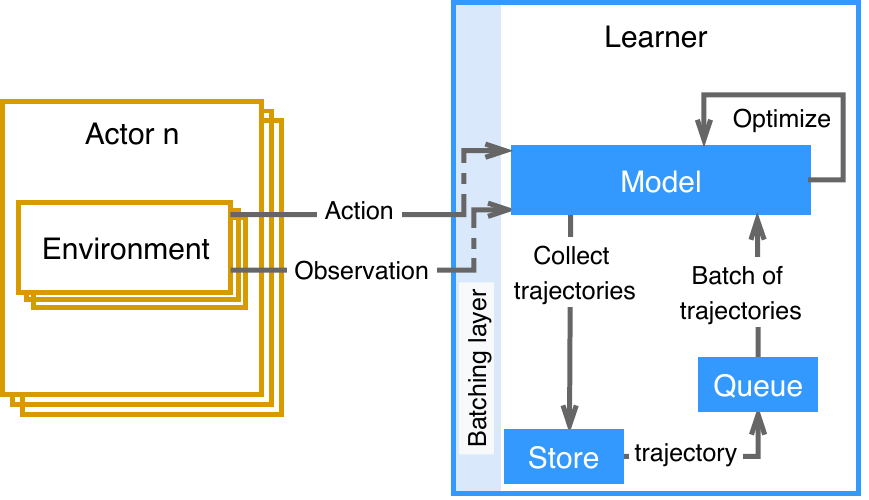}
        \caption{SEED~\cite{espeholt2020seed} 
        }
        \label{fig_seed}
    \end{subfigure}
    \caption{Parallel Deep Reinforcement Learning algorithms architectures}
    \label{fig:prallel_RL}
    \vspacecontrol
\end{figure*}
Despite the success of DRL and ESs, they are still computationally intensive approaches to tackle sequential decision-making problems. 
Parallel execution is thus an important approach to speed up the computation~\cite{nair2015massively}. Below, we look into the rich literature of parallel DRL and ES algorithms. 

\subsubsection{Parallelism in Deep Reinforcement Learning}
In parallel-DRL many agents (or actors) run in parallel to accelerate the learning process. 
Each actor gathers its own learning experiences. These experiences are, then, shared to optimize a global network (Figure~\ref{fig:prallel_RL})~\cite{Grounds2008, Dean2012}. 
The rest of this section presents important parallel-DRL algorithms. 

\noindent\textbf{Gorila}
\cite{nair2015massively} is the first massively distributed architecture for DRL. It consists of four major components:  actors,  learners, a parameter server, and a replay buffer (Figure~\ref{fig_Gorila}). Each actor has its Q-network. It interacts with an instance of the same environment and stores the generated experiences (i.e., a set of \{$s,a,r,s'$\}) in the replay buffer. 
Learners sample the experience replay buffer and use DQN to compute gradients. Sampling from a buffer reduces the correlation between data updates and the effect of non-stationarity in the data. These gradients are then sent asynchronously to the parameter server to update its Q-network. After that, the parameter server updates the actors' and learners' Q-networks to synchronize the learning process.

\noindent\textbf{A3C \& GA3C.}
While using a replay buffer helps in stabilizing the learning process, it requires additional memory and computational power and can only be used with off-policy algorithms. Motivated by these limitations, \citet{mnih2016asynchronous} introduced the Asynchronous Advantage Actor-Critic (A3C) as an alternative to Gorila. A3C consists of a global network and multiple agents each with its own network  (Figure~\ref{fig_A3C}). 
The agents are implemented as CPU threads within a single machine, which reduces the communication cost imposed by distributed systems such as Gorila. 
The agents interact in parallel with their independent copy of the environment. Each agent calculates the value and the policy gradients which are used to update the global network parameters. This method of learning diversifies and decorrelates data updates which stabilize the learning process. GA3C~\cite{babaeizadeh2017reinforcement} makes use of GPUs and shows better scalability and performance than A3C.

\noindent\textbf{Batched A2C \& DPPO.}
A downside of A3C is that asynchronous updates may lead to sub-optimal collective updates to the global network. To overcome this, 
Batched Advantage Actor-Critic (Batched A2C) employs a master node (or a coordinator) to synchronize the update process of the global network~\cite{clemente2017efficient}. Batched A2C tries to capitalize on the advantages of both Gorila and A3C. Similar to Gorila, Batched A2C runs on GPUs and the number of actors is highly scalable while still running on a single machine akin to A3C and GA3C \cite{babaeizadeh2017reinforcement}.
Figure \ref{fig_GPUpar} presents the Batched A2C architecture.  At each time step, Batched A2C samples from the policy and generates a batch of actions for $n_w$ workers on $n_e$ environment instances. The resulting experiences are then stored and used by the master to update the policy (global network). The batched approach allows for easy parallelization by synchronously updating a unique copy of the parameters, with the drawback of higher communication costs. Distributed Proximal Policy Optimization (DPPO)~\cite{heess2017emergence} features architecture similar to that of A2C, and uses the PPO~\cite{schulman2017proximal} algorithm for learning. 
\begin{figure*}[t]
    \centerline{\includegraphics[width=\textwidth]{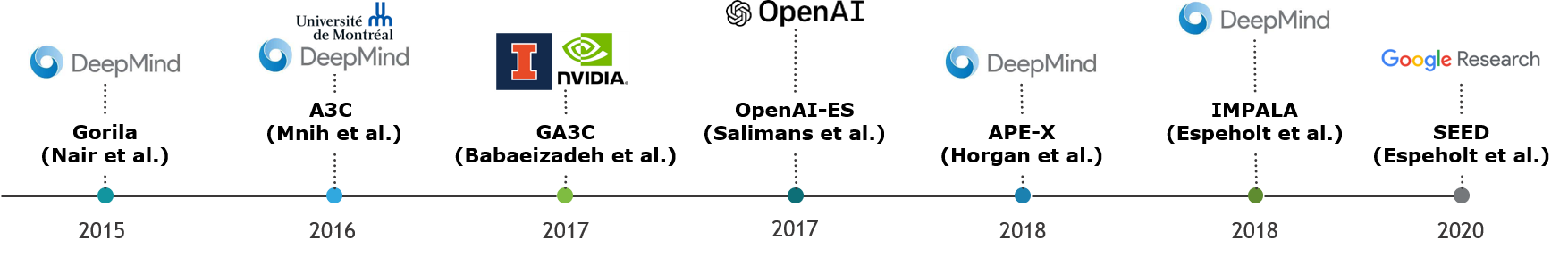}}
    \caption{Parallel Deep Reinforcement Learning and Evolution Strategies algorithms shown on a timeline}
    \label{fig:parallel-es-drl:timeline}
    \vspacecontrol
\end{figure*}

\noindent\textbf{Ape-X \& R2D2.} 
Ape-X~\cite{horgan2018distributed} extends the prioritized experience buffer to the parallel-DRL settings and shows that this approach is highly scalable. The Ape-X architecture consists of many actors, a single learner, and a prioritized replay buffer (Figure \ref{fig_apex}). Each actor interacts with its instance of the environment, gathers data, and computes its initial priorities. The generated experiences are stored in a shared prioritized buffer. The learner samples the buffer to update its network and the priorities of the experiences in the buffer. In addition, the learner also periodically updates the network parameters of the actors. Ape-X's distributed architecture can be coupled with different learning algorithms such as DQN~\cite{Mnih2015} and DDPG~\cite{lillicrap2019continuous}.
R2D2~\cite{kapturowski2018recurrent} has a similar architecture but outperforms Ape-X using recurrent neural network (RNN)-based RL agents.

\noindent\textbf{IMPALA.}
Due to the use of an on-policy method in an off-policy setting GA3C~\cite{babaeizadeh2017reinforcement} suffers from poor convergence. IMPALA~\cite{espeholt2018impala} corrected this with the use of V-trance: an off-policy actor-critic algorithm that aims at mitigating the effect of the lag between when actions are taken by the actors and when the learner estimates the gradient in distributed settings. IMPALA's architecture consists of multiple actors interacting with their environment instances (Figure~\ref{fig_impala}). However, different from  A3C's actors, IMPALA's actors send the gathered experiences (instead of the gradients) to the learner. The learner then utilizes these experiences to optimizes its policy and value functions. After that, it updates the actors' models parameters.
The separation between acting and learning and V-trace enable IMPALA to have stable learning while achieving high throughput.
When training a very deep model the speed of a single GPU is often the bottleneck. To overcome this challenge, IMPALA (in addition to a single learner) supports multiple synchronized learners.

\noindent\textbf{SEED.} 
SEED~\cite{espeholt2020seed} improves on the IMPALA system by moving inference to the learner (Figure~\ref{fig_seed}).
Consequently, the trajectories collection becomes part of the learner and the actors only send observations and actions to the learner.  SEED makes use of TUPs and GPUs and shows significant improvement over other approaches. 
\begin{table*}[t]
\centering 
    \caption{Parallelized Deep Reinforcement Learning and Evolution Strategies systems}
    \begin{tabular}{p{1.2cm}|p{4.4cm}|p{4.8cm}|p{5.2cm}|p{0.3cm}}
        \hline \textbf{Algorithms} & \textbf{Architecture} & \textbf{Experiments} & \textbf{Limitations} & \textbf{Ref.}\\
        \hline 
        Gorila & 
        replay buffer, actors, learners, and the parameter server each runs on a separate machine; GPU &
        outperforms DQN in 41/49 Atari games with reduced wall-time & 
        high bandwidth for communicating gradients and parameters  &
        \cite{nair2015massively} \cite{espeholt2018impala} \\ 
        \hline
        A3C &
        many actors each running on a CPU thread and update a global network & 
        outperforms Gorila on the Atari games while training for half the time &
        possibility of inconsistent parameter updates; large bandwidth between learner and actors; does not make use of hardware accelerators &
        \cite{mnih2016asynchronous} \cite{espeholt2018impala} \\
        \hline
        Batched A2C & 
        multi-actors, a master, which synchronizes actors' updates, and a global network; GPU & 
        requires less training time as compared to Gorila, A3C, and GA3C & 
        high variance in complex environments limits performance; episodes of varying length cause a slowdown during initialization & 
        \cite{clemente2017efficient}\\
        \hline 
        Ape-X  & 
        multi-actors, a shared learner, and prioritized replay memory; CPU/GPU & 
        outperforms Gorila and A3C on the Atari domain with less wall-clock training time & 
        inefficient CPUs usage; large bandwidth for communicating between  actors and learner & 
        \cite{horgan2018distributed}\\
        \hline
        IMPALA  & 
        multi-actors; single or multiple learners; replay buffer; GPUs&
        outperforms Batched A2C and A3C. Less sensitive to hyperparameters selection than A3C &
        uses CPUs for inference which is inefficient; requires large bandwidth for sending parameters and trajectories &
        \cite{espeholt2018impala}\\
        \hline
        SEED & multi-actors and a single learning; GPU/TPU & surpasses the performance of IMPALA & centralized inference may lead to increase latency & \cite{espeholt2020seed} \\ 
        \hline
        OpenAI ES  & 
        set of parallel workers; CPUs & 
        outperforms other solution on most Atari games with less training: better than A3C in 23 games and worse in 28 &
        evaluates many episodes requiring a lot of CPU time: 4000 CPU hours for a single ES run & 
        \cite{salimans2017evolution} \cite{chrabaszcz2018basics}\\
        \hline
    \end{tabular}
    \label{tab:par}
\end{table*}

\subsubsection{Parallelism in Evolution Strategies}
\begin{figure}[t]
    \centerline{\includegraphics[width=\columnwidth]{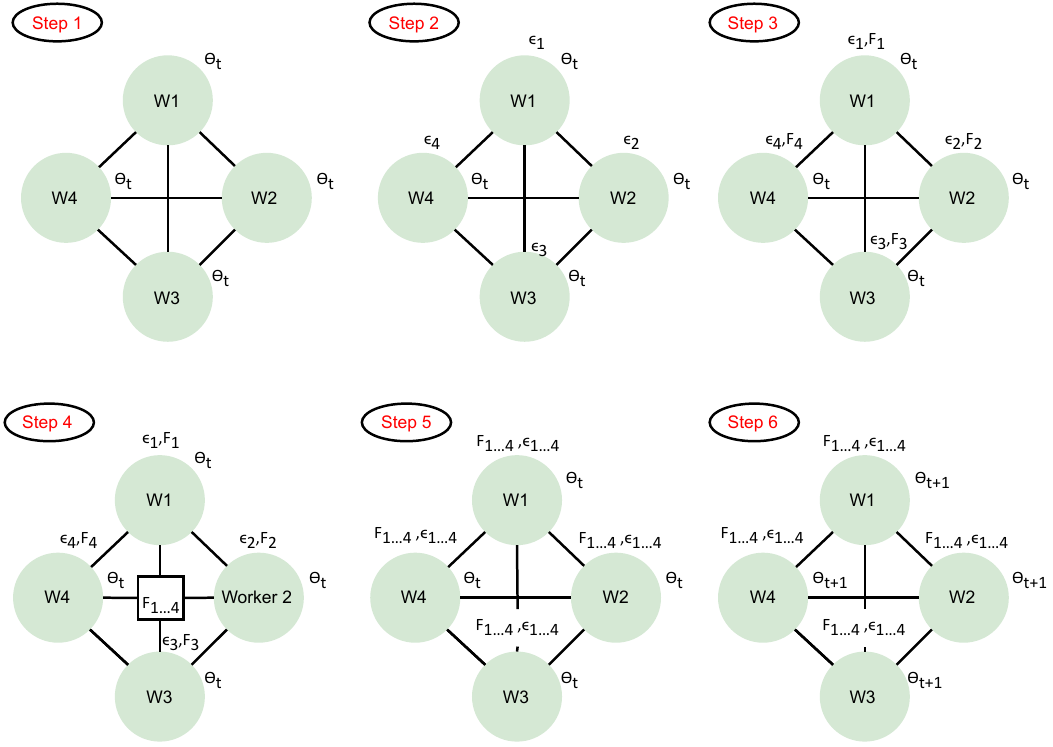}}
    \caption{Parallelization steps in OpenAI-ES~\cite{salimans2017evolution}} 
    \label{fig_parallel}
    \vspacecontrol
\end{figure}
Compared to DRL, ES algorithms require significantly less bandwidth to parallelize a given task. 
\citet{salimans2017evolution} proposed OpenAI-ES: an algorithm derived from NES (see Section~\ref{sec:fund_alg}) that directly optimizes the parameters \pmb{$\theta$} of a policy. 
By sharing the seeds of the random processes prior to the optimization process, OpenAI-ES requires exchanging only scalars (minimal bandwidth) between workers to parallelize the search process. 
The main steps of OpenAI-ES are illustrated in Figure~\ref{fig_parallel} and summarized as follows:
\begin{enumerate}
  \item Sample a Gaussian noise vector, $\varepsilon_i \sim N(0,I)$.
  \item Evaluate workers' fitness functions, $F_i\leftarrow F(\pmb{\theta}_t,\sigma\varepsilon_i)$.
  \item Exchange the fitness values, $F_i$, between the workers.
  \item Reconstruct $\varepsilon_i$ using known random seeds.
  \item Adjust parameters according to $\pmb{\theta}_{t+1} \leftarrow  \pmb{\theta}_t+\alpha\frac{1}{n\sigma}\sum_{j=1}^{n}F_j\varepsilon_j$, where $\pmb{\theta}$ is a weighted vector of a DNN.
  \item Repeat from step 2 until termination. 
\end{enumerate}
Several researchers proposed algorithms inspired by OpenAI-ES \cite{salimans2017evolution}. For example, \citet{conti2018improving} proposed Novelty Search Evolution Strategy (NS-ES) algorithm which hybridizes OpenAI-ES with Novelty Search (NS)---a directed exploration algorithm. The authors also introduced a variant of NS-ES by replacing NS with Quality Diversity (QD) algorithm. 
Their results show that the NS- and QD-based algorithms improve ES algorithms performance on RL tasks with sparse rewards, as they help avoid local optima. 
\citet{Liu2019TrustRE} proposed Trust Region Evolution Strategies (TRES). TRES is more sampled data efficient than classical ESs. It optimizes a surrogate objective function that enable reusing sampled data multiple times. 
TRES utilizes random seeds sharing introduced by \cite{salimans2017evolution} to achieve extremely low bandwidth. 
Finally, \citet{ijcai2019} proposed Evolution Strategy with Progressive Episode Lengths (PEL). The main idea of PEL is to allow an agent to do small and easy tasks to gain knowledge quickly and then to use this knowledge to tackle more complex tasks. PEL leverages the same parallelization idea as OpenAI-ES~\cite{salimans2017evolution} and shows a great improvement over canonical ES algorithms. 

\subsubsection{Comparison}
Our observations about parallelizing DRL and ES are: 
\begin{itemize}
    \item Despite the additional complexity, parallelism accelerates the execution of DRL and ES algorithms.
    \item Parallel DRL usually communicates network parameters or gradient vectors between nodes, while parallel ES share only scalar values between workers.
    \item Table \ref{tab:par} snapshots the main characteristics of the presented algorithms, and Figure~\ref{fig:parallel-es-drl:timeline} shows how parallel DRL and ES algorithms evolved over time. 
\end{itemize}

    \subsection{Exploration}
    \label{sec:exploration}
    One of the fundamental challenges that a learning agent faces when interacting with a partially known environment is the exploration-exploitation dilemma. That is, when should an agent try out suboptimal actions to improve its estimation of the optimal policy, and when should it use its current optimal policy estimation to make useful progress?  This dilemma has attracted ample attention. Below, we summarize the main exploration methods in DRL and ESs. 

\subsubsection{Exploration in (Deep) Reinforcement Learning}
Simple exploration techniques balance exploration and exploitation by selecting estimated optimal actions most of the time and random ones on occasion. This is the case for the well-known $\epsilon$-greedy exploration algorithm~\cite{sutton2018reinforcement} that acts greedily with probability $1-\epsilon$ and selects a random action with probability $\epsilon$.

More complex exploration strategies estimate the value of an exploratory action by making use of the environment-agent interaction history. 
Upper confidence bound (UCB)~\cite{auer2002finite} does that by making the reward signal equals the estimated value of a Q-function plus a value that reflects the algorithm's lack of confidence about this estimate,  
$$
r^+(s,a) = r(s,a) + B(N(s))\text{,}
$$
where $N(s)$ represents the frequency of visiting state $s$, and $B(N(s))$ is a reward bonus decreases with $N(s)$. In other words, UCB promotes the selection of actions with high rewards, $r(s,a)$, or the ones with high uncertainty (less frequently visited). The Thompson sampling method (TS)~\cite{russo2017tutorial} maintains a distribution over the parameters of a model. In the beginning, it samples parameters at random. But as the agent explores an environment, TS adapts the distribution to favor more promising parameter sets. As such, UCB and TS naturally reduce the probability of selecting exploratory actions and become more confident about the optimal policy over time. Therefore, they are inherently more efficient than $\epsilon$-greedy.

\noindent\textbf{From RL to DRL.}
DRL agents act on environments with continuous or high-dimensional state-action spaces (e.g., Montezuma's Revenge, StarCraft II). Such spaces render count-based algorithms (e.g., UCB) and the ones that require maintaining a distribution over state-action spaces (e.g., TS) useless in their original formulation. To explore such challenging environments with sparse reward signals, many algorithms have been proposed.  Generally, these algorithms couple approximation techniques with exploration algorithms proposed for simple RL settings~\cite{achiam2017surprisebased,pathak2019selfsupervised,shyam2019modelbased,kim2019emi}. Below we outline important DRL exploration algorithms. 

\textit{pseudo-count methods.} 
To extend count-based exploration methods (e.g., UCB) to DRL settings, \citet{bellemare2016unifying} approximate the counting process using a Context Tree Switching (CTS) density model. The model's goal is to provide a score that increases when a state is revisited. The score is then used to generate a reward bonus that is inversely proportional to the score value. This bonus is then added to the reward signal provided by the environment to incentive the agent to visit less-visited states.  \citet{ostrovski2017countbased} improved this approach by replacing the simple CTS density model with a neural density model called PixelCNN.
Another approach to utilize counting to explore environments with high-dimensional spaces is by mapping the observed states to a hashing table~\cite{tang2017exploration} and counting the hashing codes instead of states. Then a reward bonus similar to that of UCB is designed utilizing the hash code counts. 

\textit{Approximate posterior sampling.} 
Inspired by TS, \citet{osband2016deep} introduced Bootstrapped DQN. Bootstrapped DQN trains a DNN with N bootstrapped heads to approximate a distribution over Q-functions (bootstrapping is the process of approximating a distribution by sampling with replacement from a population multiple times and then aggregating these samples). At the start of each episode, Bootstrapped DQN draws a sample at random from the ensemble Q-functions and acts greedily with respect to this sample. This strategy enables an RL agent to do temporally extended exploration (or deep exploration) which is particularly important when the agent receives a sparse environmental reward.  \citet{chen2017ucb} integrates UCB with Bootstrapped DQN by calculating the mean and variance of a subset of the ensemble Q-functions. \citet{o2018uncertainty} combined TS with uncertainty Bellman equations to propagate the uncertainty in the Q-values over multiple timesteps.
\begin{table*}[t]
\centering
    \caption{Deep Reinforcement Learning exploration algorithms}
    \begin{tabular}{p{2cm}|p{6.1cm}|p{8cm}|p{0.3cm}}
         \hline 
          \textbf{Algorithm} & \textbf{Description} & \textbf{Experiments} & \textbf{Ref.}\\
           \hline
              Bootstrapped DQN &
              uses DNNs and ensemble Q-functions to explore an environment &
              outperforms DQN by orders of magnitude in terms of cumulative rewards &
               \cite{osband2016deep} \\
          \hline
                UCB$+$InfoGain &
                integrates  UCB  with  Q-function ensemble &
                outperforms bootstrapped DQN &
                \cite{o2018uncertainty} \\
          \hline 
                State pseudo-count &
                uses density models and pseudo-count to approximate state visitation count which is used to compute the reward bonus   &
                superior to DQN, especially in hard-to-explore environments &
                \cite{bellemare2016unifying}\\
          \hline
          VIME & 
              measures information gain as KL divergence between current and updated distribution after an observation & 
              improves the performance of TRPO~\cite{schulman2015trust}, REINFORCE~\cite{WilliamsREINFORCE1992} when added to them & 
              \cite{houthooft2017vime} \\
          \hline
          ICM & 
              uses a forward dynamic model to predict states and measures information gain as the  difference between the predicted and observed state& 
              outperforms TRPO-VIME in VizDoom (a sparse 3D environment) & 
              \cite{pathak2017curiositydriven}\\ 
           \hline 
             Episodic curiosity & 
             uses episodic memory to form the novelty bonus & 
             outperforms ICM in visually rich 3D environments from VizDoom and DMLab & 
             \cite{savinov2019episodic}\\ 
            \hline 
        Go-Explore & The previously visited states are stored in memory. In phase one Go-explore explores until a solution is found. In phase two Go-explore Robustifies the found solution & Performance improvements on hard exploration problems over other methods such as DQN+PixelCNN, DQN+CTS, BASS-hash &  \cite{ecoffet2020goexplore}\\ 
        \hline 
          Never Give Up &  combines both episodic and life-long novelties & 
          obtains a median human normalized score of 1344\%; the first algorithm that achieves non-zero rewards in the game of Pitfall & \cite{badia2020up}\\ 
          Agent57 & uses a meta-controller for adaptively selecting the right policy: ranging from purely exploratory to purely exploitative & first DRL agent that surpasses the standard human benchmark on all 57 Atari games & \cite{badia2020agent57} \\
          \hline
    \end{tabular}
    \label{tab:drlexp}
\end{table*}

\textit{Information gain.} 
In exploration based on information gain, the algorithm provides a reward bonus proportional to the information obtained after taking an action.  This reward bonus is then added to the reward provided by the environment to push the agent to explore novel (or less known) states \cite{schmidhuber2010formal}. 
\citet{houthooft2017vime} proposed to learn a transition dynamic model with a Bayesian neural network. 
The information gain is measured as the KL divergence between the current and updated parameter distribution after a new observation. Based on this information the reward signal is augmented with a bonus. 
\citet{pathak2017curiositydriven} used a forward dynamic model to predict the next state.
The reward bonus is then set to be proportional to the error between the predicted and observed next state.
To make this method effective, the authors utilized an inverse model, removing irrelevant -for the comparison- state features.
\citet{burda2018exploration} defines the exploration bonus based on the error of a neural network in predicting features of the observations given by a fixed randomly initialized neural network.

\textit{Memory-based.} 
\citet{savinov2019episodic} proposed a new curiosity method that uses episodic memory to form the novelty bonus. The bonus is computed by comparing the current observation with the observations in memory and a reward is given for observations that require some effort to be reached (effort is materialized by the number of environment steps taken to reach an observation). 
\citet{ecoffet2020goexplore} introduced Go-explore: an RL agent that aims to solve hard exploration problems such as Montezuma's Revenge and Pitfall. Go-explore runs in two phases. In phase one, the agent explore randomly, remembers interesting states and continues (after reset) random exploration from one of the interesting states (the authors assume the agent can deterministically go back to an interesting state). After finding a solution to the problem, phase two begins where the Go-explore agent robustifies its the best found solution by randomizing the environment and running imitation learning using the best solution. 
\citet{badia2020up} proposed "Never give up" (NGU): an agent that also targets hard exploration problems. NGU augments the environmental reward with a combination of two intrinsic novelty rewards:  (i) An episodic reward, which enables the agent to quickly adapt within an episode, and, (ii) the life-long novelty reward, which down-modulates states that become familiar across many episodes. Further, NGU uses a Universal Value Function Approximator (UVFA) to learn several exploration policies with different exploration-exploitation trade-offs at the same time. 
Agent57 \cite{badia2020agent57} aims to manage the tradeoff between exploration and exploitation using a "meta-controller" that adaptively selects a correct policy (ranging from very exploratory to purely exploitative) for the training phase. Agent57 outperforms the standard human benchmark on all 57 Atari games. 

\begin{figure*}[t]
    \centerline{\includegraphics[width=\textwidth]{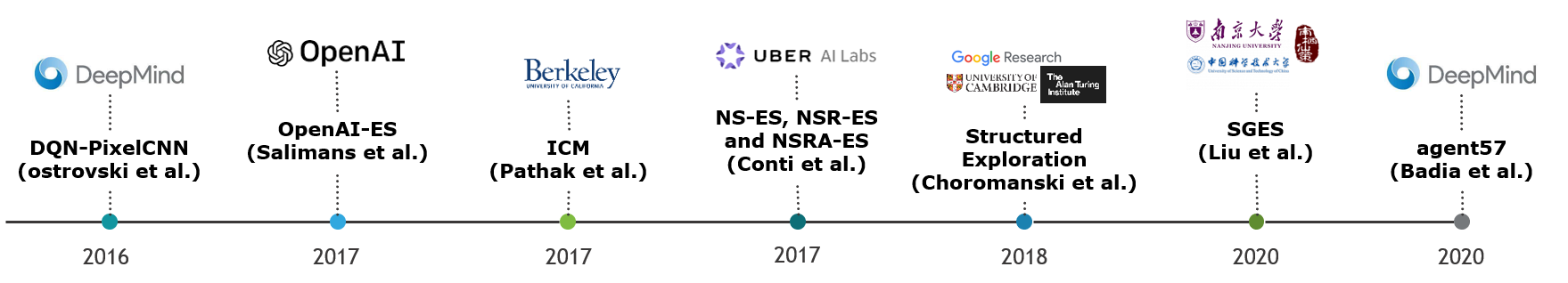}}
    \caption{Deep Reinforcement Learning and Evolution Strategies exploration algorithms shown on a timeline}
    \label{fig_exploration_history}
    \vspacecontrol
\end{figure*}

\subsubsection{Exploration in Evolution Strategies}
ES algorithms optimize the fitness score while exploring around the best solutions found so far.
The exploration is realized through the  recombination and mutation steps.  
Despite their effectiveness in exploration, ESs may still get trapped in local optima~\cite{Liu2019TrustRE, zhang2020accelerating}. To overcome this limitation, many ESs algorithms with enhanced exploration techniques have been proposed.

One way to extract approximate gradients from a non-smooth objective function, $F(\pmb{\theta})$, is by adding noise to its parameter vector, $\pmb{\theta}$. This yields a \textit{new differentiable} function, $F_{ES}(\pmb{\theta})$. OpenAI-ES~\cite{salimans2017evolution} exploits this idea by sampling noise from a Gaussian distribution and adding it to the parameter vector $\pmb{\theta}$. The algorithm then optimizes using stochastic gradient ascent. 
Additionally, OpenAI-ES relays on a few auxiliary techniques to enhance its performance: virtual batch normalization \cite{salimans2016improved} for enhanced exploration, antithetic sampling~\cite{geweke1988antithetic} for reduced variance, and fitness shaping~\cite{wierstra2014natural} for improving local optima avoidance.

\citet{choromanski2018structured} proposed two strategies to enhance the exploration of Derivative Free Optimization (DFO) methods such as OpenAI-ES \cite{salimans2017evolution}: (i) structured exploration, where the authors showed that random orthogonal and Quasi Monte Carlo finite difference directions are much more effective than random Gaussian directions for parameter exploration; and (ii) compact policies, whereby imposing a parameter sharing structure on the policy architecture, they were able to significantly reduce the dimensionality of the problem without losing accuracy and thus speeding up the learning process.

\citet{maheswaranathan2019guided} proposed Guided ES: a random search that is augmented using surrogate gradients which are correlated with the true gradient. The key idea is to track a low-dimensional subspace that is defined by the recent history of surrogate gradients. Sampling this subspace leads to a drastic reduction in the variance of the search direction. However, this approach has two shortcomings: (i) the bias of the surrogate gradients needs to be known; and (ii) when the bias is too small, Guided ES cannot find a better descent direction than the surrogate gradient. \citet{meier2019improving} draw inspiration from how momentum is used for optimizing DNNs to improve upon Guided ES~\cite{maheswaranathan2019guided}. The authors showed how to optimally combine the surrogate gradient directions with random search directions and how to iteratively approach the true gradient for linear functions. They assessed their algorithm against a standard ESs algorithm on different tasks showing its superiority.

\citet{choromanski2019complexity} noted that fixing the dimensionality of subspaces (as in Guided ES~\cite{maheswaranathan2019guided}) leads to suboptimal performance. Therefore, they proposed ASEBO: an algorithm that adaptively controls the dimensionality of subspaces based on gradient estimators from previous iterations. ASEBO was compared to several ESs and DRL algorithms and showed promising averaged performance.

\citet{Liu_2020HistoricalES} proposed Self-Guided Evolution Strategies (SGES). This work is inspired by both ASEBO \cite{choromanski2019complexity} and Guided ES \cite{maheswaranathan2019guided}. Further,  
it is based on two main ideas: leveraging historical estimated gradients and building a guiding subspace from which search directions are sampled probabilistically. The results show that SGES outperforms Open-AI~\cite{salimans2017evolution}, Guided ES~\cite{maheswaranathan2019guided}, CMA-ES and vanilla ES.

The aforementioned methods suffer from the curse of dimensionality due to the high variance of Monte Carlo gradient estimators.
Motivated by this, \citet{zhang2020novel} proposed Directional Gaussian Smoothing Evolution Strategy (DGS-ES). It encourages non-local exploration and improves high-dimensional exploration. In contrast to regular Gaussian smoothing, directional Gaussian smoothing conducts $1D$ non-local explorations along $d$ orthogonal directions. The Gauss-Hermite quadrature is then used for improving the convergence speed of the algorithm. Its superior performance is showcased by comparing it to many algorithms including OpenAI-ES \cite{salimans2017evolution} and ASEBO~\cite{choromanski2019complexity}.

To encourage exploration in environments with sparse or deceptive reward signals, \citet{conti2018improving} proposed hybridizing ESs with directed exploration methods (i.e., Novelty Search (NS)  \cite{Lehman2011NS} and Quality Diversity (QD) \cite{Pugh2016QD}). 
The combination resulted in three algorithms: NS-ES, NSR-ES, and NSRA-ES. 
NS-ES builds on the OpenAI-ES exploration strategy. OpenAI-ES approximates a gradient and takes a step in that direction. In NS-ES, the gradient estimate is that of the expected novelty. It gives directions on how to change the current policy’s parameters $\pmb{\theta}$ to increase the average novelty of the parameter distribution.
NSR-ES is a variant of NS-ES. It combines both the reward and novelty signals to produce policies that are both novel and high-performing. 
NSRA-ES is an extension of NSR-ES that dynamically adapts the weights of the novelty and the reward gradients for more optimal performance. 

\begin{table*}[t]
\centering
    \caption{Evolution Strategies exploration algorithms}
    \begin{tabular}{p{2cm}|p{6.1cm}|p{8cm}|p{0.3cm}}
         \hline 
          \textbf{Algorithm} & \textbf{Description} & \textbf{Experiments} & \textbf{Ref.}\\
          \hline
          OpenAI-ES & adds Gaussian noise to the parameter vector, computes a gradient, and takes a step in its direction & 
          improves exploratory behaviors as compared to TRPO on tasks such as learning gaits of the MuJoCo humanoid walker & 
          \cite{salimans2017evolution} \\
          \hline
          Structured Exploration &
          complements OpenAI-ES~\cite{salimans2017evolution} with  structured exploration and compact policies for efficient exploration& 
          solves robotics tasks from OpenAI Gym using NN with 300 parameters (13x fewer than OpenAI-ES) and with near linear time complexity& 
          \cite{choromanski2018structured}\\
          \hline
          Guided ES & 
          leverages surrogate gradients to define a low-dimensional subspace for efficient sampling & 
          improves over vanilla ESs and first-order methods that directly follow the surrogate gradient &
          \cite{maheswaranathan2019guided}\\
          \hline
          ASEBO & 
          adapts the dimensionality of the subspaces on-the-fly for efficient exploration & 
          optimizes high-dimensional balck-box functions and performs consistently well across several tasks compared to  state-of-the-art algorithms & 
          \cite{choromanski2019complexity}\\
          \hline
          DGS-ES & 
          uses directional Gaussian smoothing to explore along non-local orthogonal directions. It leverages Guss-Hermite quadrature for fast convergence. & 
          improves on state-of-the-art algorithms (e.g., OpenAI-ES
           and ASEBO) on some problems &
           \cite{zhang2020novel}\\
          \hline
          Iterative gradient estimation refinement 
          & 
          iteratively uses the last update direction as a surrogate gradient for the gradient estimator. Over time this will result in improved gradient estimates. & 
          converges relatively fast to the true gradient for linear functions. It improves gradient estimation of ESs at no extra computational cost on MNIST and RL tasks &\cite{meier2019improving} \\
          \hline
          SGES & 
          adapts a low-dimensional subspace on the fly for more efficient sampling and exploring & 
          has lower gradient estimation variance as compared to OpenAI-ES. Superior performance over ESs algorithms such as OpenAI-ES, Guided ES, ASEBO, CMA-ES on blackbox functions and RL tasks & 
          \cite{Liu_2020HistoricalES}\\ 
          \hline
         NS-ES, NSR-ES, and NSRA-ES & Hybridize Novelty search (NS) and quality diversity (QD) algorithms with ESs to improve the performance of ESs on sparse RL problems. & 
         avoid local optima encountered by ESs while achieving higher performance on Atari and simulated robot tasks & \cite{conti2018improving}\\
          \hline
    \end{tabular}
    \label{tab:es}
\end{table*}

\subsubsection{Comparison}

Our observations of this section are summarized below.
\begin{itemize}
    \item The exploration-exploitation dilemma is still an active field of research and environments with sparse and deceptive reward signals require more sophisticated and capable exploration algorithms.
    \item Benchmarking exploration strategies happens almost exclusively in simulated/gaming environments. Consequently, the efficacy of these algorithms in real-world applications is mostly unknown. 
    \item Thanks to the recombination and mutation, ESs algorithms might suffer less from local optima than DRL ones.
    \item ESs still face some problems related to sample efficiency when exploring, as high dimensional optimization tasks can lead to high variance gradients estimates.
    \item Table~\ref{tab:drlexp} and Table~\ref{tab:es} summarize some important characteristics of  DRL and ESs exploration algorithms.
\end{itemize}

    \subsection{Non-Markov settings}
    \label{sec:dynamic}
    The Markov property denotes the situation where the future states of a process depend only on the current state and not on events or states from the past.
The degree to which agents can observe (changes in) the environment has an impact on their decision behavior. In certain favorable scenarios the state of the agent in its environment might be fully observable (e.g., using sensors) to an extent such that the Markov assumption holds. In other cases, the state of the environment is only partially observable and/or the agent faces a distribution of environments (Meta-RL).

\subsubsection{Partially Observable}
In many real-world applications, agents can only partially observe the state of their environments and might only have access to their local observations. This means agents need to take into accent the history of observations---actions and rewards---to produce a better estimation of the underlying hidden state \cite{mccallum1997reinforcement,ortner2014selecting,franccois2019overfitting}. These problems are usually modeled as a partially observable Markov decision process (POMDP). Researchers have addressed the POMDP problem setup through the proposal of many RL models and evolutionary strategies. In DRL, one possibility is to employ a neural network with a recurrent architecture that enables agents to consider past observations \cite{MarlHausknecht,Igel2004Neuroevolution}.

\subsubsection{Meta Reinforcement Learning}
Meta-RL is concerned with learning a policy that can be quickly generalized across a distribution of tasks or environments (modeled as MDPs). 
Generally, a meta-learner achieves that through two stages optimization process: first, a meta-policy is trained on a distribution of similar tasks with the hope of learning the common dynamics across these tasks; then, the second stage fine-tunes the meta-policy while acting on a particular task sampled from a similar but unseen task distribution \cite{Schaul:2010}. 
Examples of meta-RL tasks include: navigating towards distinct goals \cite{gupta2018metareinforcement}, going through different mazes \cite{mishra2018simple}, dealing with component failures~\cite{nagabandi2019learning}, or driving different cars \cite{garcia2019metamdp}. 

Meta-RL can be subdivided into two categories \cite{gupta2018metareinforcement}: RNN-based \cite{wang2017learning,duan2016rl2} and gradient-based learners \cite{finn2017modelagnostic,li2017metasgd}.

\noindent\textbf{Recurrent Models (RNN-based learners).}
Leveraging the agent-environment interaction history provides more information, which leads to improved learning \cite{garcia2019metamdp,hochreiter2001learning}. 
This idea can be implemented using Recurrent Neural Networks (RNNs) (or other recurrent models) \cite{santoro2016oneshot,duan2016rl2,wang2017learning,mishra2018simple}. The RNNs can be trained on a set of tasks to learn a hidden state (meta-policy), then this hidden state can be further adapted given new observations from an unseen task. 

General architecture of a meta-RL algorithm is illustrated in Figure~\ref{fig_metaRL} \cite{Botvinick2019}, where an agent is modeled as two loops, both implementing RL algorithms. The outer loop samples a new environment in every iteration and tunes the parameters of the inner loop. Consequently, the inner loop can adjust more rapidly to new tasks by interacting with the associated environments and optimizing for maximal rewards. 

\citet{duan2016rl2} and \citet{wang2017learning} proposed analogous recurrent Meta-RL agents: $R^2$ and DRL-meta, respectively. They implemented a long-short term memory (LSTM) and a gate recurrent unit (GRU) architecture in which the hidden states serve as a memory for tracking characteristics of interaction trajectories. 
The main difference between both approaches relates to the set of environments. Environments in \cite{wang2017learning} are issued from a parameterized distribution \cite{robles2019learning}. In contrast, those in \cite{duan2016rl2} are relatively unrelated \cite{robles2019learning}.

Such RNN-based methods have proven to be efficient on many RL tasks. However, their performance decreases 
as the complexity of the task increases, especially with long temporal dependencies. 
Additionally, short-term memory is challenging for RNN due to the vanishing gradient problem. Furthermore, RNN-based meta-learners cannot pinpoint specific prior experiences \cite{mishra2018simple,huisman2020survey}.

To overcome these limitations, \citet{mishra2018simple} proposed Simple Neural Attentive Learner (SNAIL). It combines temporal convolutions and attention mechanisms. The former aggregates information from past experiences and the latter pinpoints specific pieces of information. 
SNAIL's architecture consists of three main parts: (i) DenseBlock, a causal 1D-convolution with specific dilation rate; (ii) TCBlock, a series of DenseBlocks with exponentially increasing dilation rates; and (iii) AttentionBlock, where key-vlaue lookups take place. This general-purpose model has shown its efficacy on tasks ranging from supervised to reinforcement learning. Despite that, challenges such as the long time needed for getting the right architectures of TCBlocks and DenseBlocks. \cite{huisman2020survey} persist.

\noindent\textbf{Gradient-Based Models.}
Model Agnostic Meta-Learning (MAML)~\cite{finn2017modelagnostic} realizes meta-learning principles by learning an initial set of parameters, $\pmb{\theta_0}$, of a model such that taking a few gradient steps is sufficient to tailor this model to a specific task. More precisely, MAML learns $\pmb{\theta_0}$ such that for any randomly sampled task, $\mathcal{T}$, with a loss function, $\mathcal{L}$, the agent will have a modest loss after $n$ updates: 
$$
 \pmb{\theta_0} = \argmin\limits_{\pmb{\theta}} \mathbb{E_{\mathcal{T}}}\bigg[ \mathcal{L}_\mathcal{T}\bigg(U_\mathcal{T}^n(\pmb{\theta}) \bigg) \bigg]
$$
where $U_\mathcal{T}^n(\pmb{\theta})$ refers to an update rule such as gradient descent. 

\citet{nichol2018firstorder} proposed Reptile a first-order meta-learning framework, that is considered to be an approximation of MAML. Similar to first-order MAML (FOMAML), Reptile does not calculate second derivatives, which makes it less computationally demanding. It starts by repeatedly sampling a task, then performing $N$ iterations of stochastic gradient descent (SGD) on each task to compute a new set of parameters. Then, it moves the model weights towards the new parameters. Next, we look at how meta-learning tries to make ESs more efficient.

\begin{figure}[t] 
\centerline{\includegraphics[width=\columnwidth]{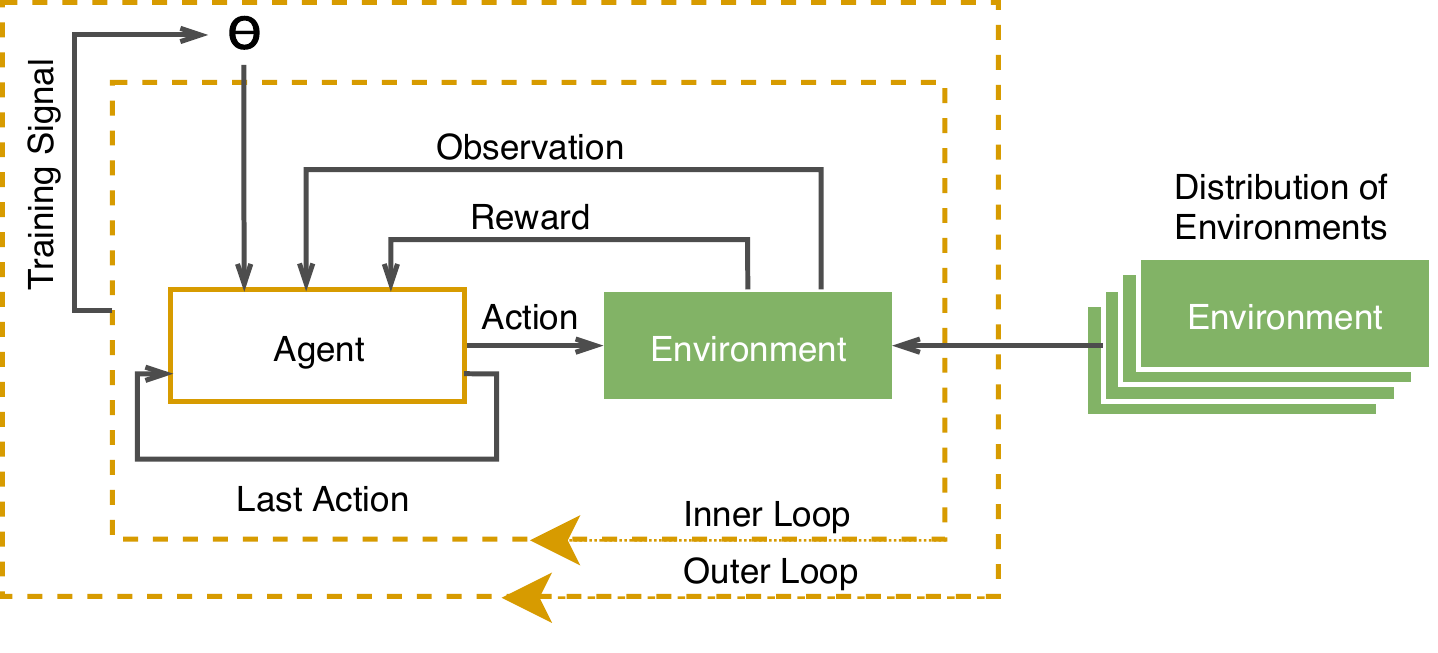}}
\caption{Schematic of Meta-reinforcement Learning;  
illustrating the inner and outer loops of training \cite{Botvinick2019}}
\label{fig_metaRL}
\vspacecontrol
\end{figure}

\begin{table*}[t]
\centering
    \caption{Gradient-based Meta Reinforcement Learning.}
    \begin{tabular}{p{2.5cm}|p{6cm}|p{7.5cm}|p{0.5cm}}
         \hline  
          \textbf{Algorithms} & \textbf{Description} & \textbf{Experiments} & \textbf{Ref.}\\
          \hline
           DRL-meta & 
           trains an RNN on a distribution of RL tasks. The RNN serves as a dynamic task embedding storage.  DRL-meta uses the LSTM architecture & 
           outperforms other benchmarks on the bandits' problems; properly adapts to invariances in MDP tasks & \cite{wang2017learning} \\
          \hline
          $RL^2$ & 
          trains an RNN on a distribution of RL tasks. The RNN serves as a dynamic task embedding storage.  $R^2$ uses the GRU architecture & 
          comparable to theoretically optimal algorithms in small-scale  settings. It has the potential to scale to high-dimensional tasks &  
          \cite{duan2016rl2}\\
          \hline
          SNAIL & combines temporal convolution and attention mechanisms & outperforms LSTM and MAML &  \cite{mishra2018simple}\\
          \hline
          MAML & 
          given a task distribution, it searches for optimal initial parameters such that a few gradient steps are sufficient to solve tasks drawn from that distribution. &
          outperforms classical methods such as random and pretrained methods & 
          \cite{finn2017modelagnostic}\\ 
          \hline
          Reptile &
          similar to first-order MAML& 
          on-par with the performance of MAML& \cite{nichol2018firstorder}\\
          \hline
    \end{tabular}
    \label{tab:rlDynamic}
\end{table*}

\subsubsection{Meta Evolution Strategies}

\citet{gajewski2019evolvability} introduced ``Evolvability ES'', an ES-based meta-learning algorithm for RL tasks. It combines concepts from evolvability search \cite{Mengistu2016}, ESs \cite{salimans2017evolution}, and MAML \cite{finn2017modelagnostic} to encourage searching for individuals whose immediate offsprings show signs of behavioral diversity (that is, it searches for parameter vectors whose perturbations lead to differing behaviors) \cite{Mengistu2016}.  Consequently, Evolvability ES facilitates adaptation and generalization while leveraging the scalability of ESs \cite{gajewski2019evolvability,katona2021quality}. Evolvability ES shows a competitive performance to gradient-based meta-learning algorithms. 
Quality Evolvability ES~\cite{katona2021quality} noted that the original Evolvability ES~\cite{hospedales2020metalearning} can only be used to solve problems where the task performance and evolability align. To eliminate this restriction, Quality Evolvability ES optimizes for both -task performance and evolability- simultaneously.

\citet{song2020esmaml} argue that policy gradient-based Model Agnostic Meta Learning (MAML) algorithms~\cite{finn2017modelagnostic} face significant difficulties when estimating second derivative using backpropagation on stochastic policies. Therefore, they introduced ES-MAML, a meta-learner that leverages ES~\cite{salimans2017evolution} for solving MAML problems without estimating second derivatives. The authors empirically showed that ES-MAML is competitive with other Meta-RL algorithms.  
\citet{song2020rapidly} combined Hill-Climbing adaptation with ES-MAML to develop noise-tolerant meta-RL learner. The authors showcased the performance of their algorithm using a physical legged robot.

\citet{wang2020instance} incorporated an instance weighting mechanism with ESs to generate an adaptable and salable  meta-learner,  Instance Weighted Incremental Evolution Strategies (IW-IES). 

\citet{wang2020instance} introduced Instance Weighted Incremental Evolution Strategies (IW-IES). It incorporates an instance weighting mechanism with ESs to generate an adaptable and salable meta-learner.
IW-IES assigns weights to offsprings proportional to the amount of new knowledge they acquire. The weights are assigned based on one of the two metrics: instance novelty and instance quality. Compared to ES-MAML, IW-IES proved competitive for robot navigation tasks.

Meta-RL is particularly suited for tackling the sim-to-real problem: simulation provides previous experiences that are used to learn a general policy, and the data obtained from operating in the real world fine-tunes that policy~\cite{tan2018simtoreal}.
Examples of using Meta-RL to train physical robots include: \citet{nagabandi2019learning} built on top of MAML a model-based meta-RL agent to train a legged millirobot; \citet{arndt2019meta} proposed a similar framework to MAML to train a robot on a task of hitting a hockey puck; and \citet{song2020rapidly} introduced a variant of ES-MAML to train and quickly adapt the policy commanding a legged robot.

\begin{table*}[t]
\centering
    \caption{Evolution Strategies for Meta Reinforcement learning}
    \begin{tabular}{p{2.5cm}|p{6cm}|p{7.5cm}|p{0.5cm}}
         \hline 
          & \textbf{Description} & \textbf{Experiments} & \textbf{ref.}\\  
          \hline
          Evolvability ES 
          & combines concepts from evolvability search, ES, and MAML to enable a quickly adaptable meta-learner & 
          competitive with MAML on 2D and 3D locomotion tasks & 
          \cite{gajewski2019evolvability}\\
          \hline 
          ES-MAML & 
          uses ESs to overcome the limitations of MAML & 
          competitive with policy gradient methods; yields better adaptation with fewer queries & \cite{song2020esmaml}\\
          \hline 
          IW-IES &  uses NES for updating the RL policy network parameters in a dynamic environment &
          outperforms ES-MAML on set of robot navigation tasks & 
          \cite{wang2020instance}\\
          \hline
    \end{tabular}
    \label{tab:esDynamic}
\end{table*}

\subsubsection{Comparison}
Key observations of this section can be summarized as:
\begin{itemize}
    \item In many cases, RL and ESs are faced with problems that are more complex than the traditional MDP setting, such as in the partially observable case and the meta-learning setting.
    \item Meta-learning enables an agent to explore more intelligently and acquire useful knowledge more quickly. 
    \item There are two main approaches for Meta-RL: gradient-based and recurrent models. Gradient-based Meta-RL is generally a two-stage optimization process: first, it optimizes on a task distribution level, and then, fine-tunes for a specific task. Meta-RL with recurrent models make use of specific recurrent architectures to learn how to act in a distribution of environments.
    \item There are many challenges in Meta-RL methods, such as estimating first and second-order derivatives, high variance, and high computation needs.
    \item ES-based meta-RL attempts to address the limitations of gradient-based Meta-RL; however, ES-based meta-RL itself faces a different set of challenges such as the sample efficiency. 
   \item Meta-RL is particularly suited for tackling the sim-to-real problem. For instance, a generic policy is trained in simulation and fine-tuned via the interaction with real world. 
\end{itemize}

    \subsection{Learning in multiagent settings}
    \label{sec:multiagent}
    \begin{figure}[t]
\centerline{\includegraphics[width=\columnwidth]{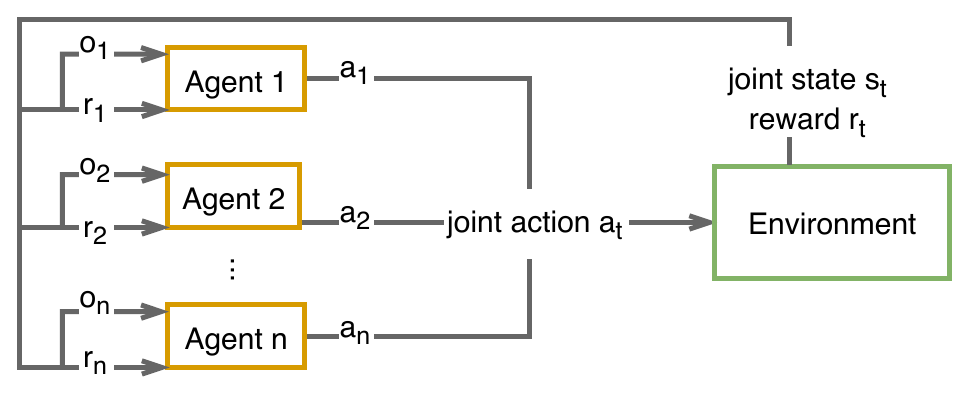}}
\caption{Multi-agent Reinforcement Learning Overview \cite{Nowe2012}}
\label{fig_MARL}
\vspacecontrol
\end{figure}


A multi-agent system (MAS) is a distributed system of multiple cooperating or competing (physical or virtual) agents, working towards maximizing their own objectives within a shared environment \cite{MarlHu}. 
Currently, MAS form one of the leading research areas of Artificial Intelligence due to their wide applicability. Virtually any application that can be partitioned and parallelized can benefit from using multiple agents.

\subsubsection{multi-agent Reinforcement Learning}
An MAS can be combined with DRL to form a Multi-agent Deep Reinforcement Learning (MADRL) system which addresses sequential decision-making problems for multiple agents sharing a common environment (Figure \ref{fig_MARL}). MADRL agents are trained to learn certain behaviors through interaction with the environment and optionally with other agents. 
%
Since the environment and the reward states are affected by the joint actions of all agents, the single-agent MDP model cannot be directly applied to MADRL systems, as they do not adhere to the Markov property.
The Markov (or Stochastic) games (MG)~\cite{shapley1953stochastic} framework comes as a generalization of the MDP that captures the entanglement of the multiple agents.  
There are several important properties to be considered when considering MADRL systems. In the following section we will discuss each of these properties and their resulting impact on the overall system. 

\noindent\textbf{Setup: Cooperative vs Competitive.} In a cooperative game, also known as a team problem, the agents seek to maximize a common reward signal by taking actions that favor their outcome, while taking into account their effects on other agents. Most contemporary applications are based upon a cooperative setup. Examples of this scenario include foraging, exploration and warehouse robots. 

One of the main challenges of learning in a cooperative setting is termed as \textit{multi-agent credit assignment problem} which refers to how to divide a reward obtained on a team level amongst individual learners \cite{panait2005cooperative}.  Due to the complex interaction dynamics of the agents, it is not trivial to determine whose actions were beneficial to the group reward.


On the other hands, agents in a competitive game receive different reward signals based on the overall outcome of the joint actions. In this setup, certain actions might be beneficial to one set of agents while being indifferent or disadvantageous for the other agents.

\noindent\textbf{Control: Centralized vs Decentralized.}
Another important distinction to make for MADRL systems is the centralized versus decentralized control approach. In the case of centralized control, there exists a single control entity that governs the decisions of all agents based on all available joint actions, joint rewards and joint observations. While this approach enables optimal decisions, it quickly becomes computationally intractable as the number of agents in a system grows. Additionally, this creates the risk of a single point of failure since the whole system could fail if the central controller breaks.

The decentralized approach does not make use of a central controller and relies on agents to make decisions independently, based on the information available to them locally. Decentralized systems can be subdivided into two categories: "A decentralized setting with networked agents", and ''A fully decentralized setting`` \cite{zhang2019multiagent}. The former setup involves agents which can communicate with other agents and use the shared information to optimize their actions. In the latter scenario, agents make independent decisions without information exchange. While this means that no explicit messages can be sent, it is still possible to influence the behavior of other agents by affecting their reward as seen in \cite{PRIMAL}. While the decentralized approach can provide more scalability and robustness, it also significantly increases the complexity of the system as there is no central entity that has knowledge of and can control the state of each robot. An interesting future research direction might be semi-centralized MADRL systems in which one or more central entities possess partial information of a set of agents. Alternatively, it is possible to alternate techniques between different phases of the design. \citet{MarlChen} proposed a system with centralized training and exploration and decentralized execution which can increase inter-agent collaboration and sample efficiency. 

\noindent\textbf{Challenges in Multi-agent Reinforcement Learning.}
Moving from a single-agent to a multi-agent environment brings about new complex challenges with respect to learning and evaluating outcomes. This can be attributed to several factors, including the exponential growth of the search space and the non-stationarity of the environment \cite{Lee_2020}.
Next, MADRL challenges are discussed. 

\textit{Non-stationarity}: In a MADRL system, agents are learning concurrently and their actions reshape their shared surroundings repeatedly, resulting in a non-stationay environment.
Consequently, the convergence of well-known algorithms such as Q-learning can no longer be guaranteed as the Markov property assumption of the environment is violated~\cite{hernandez2017survey,tan1993multi,Mnih2015}.
Many papers in the literature that attempt to address the non-stationarity problem. \citet{castaneda2016deep} proposed two algorithms: Deep loosely coupled Q-network (DLCQN) and deep repeated update Q-network (DRUQN). DLCQN modifies an independence degree for each agent based on the agent's negative rewards and observations. The agent then utilizes this independence degree to decide when to act independently or cooperatively. DRUQN tries to avoid policy bias by making the value of an action inversely proportional to the probability of selecting that action.
The use of an experience replay buffer with DQN enables efficient learning. However, due to the non-stationarity of the environment in MADRL settings data stored in an experience replay buffer can become outdated. To counter this unwanted behavior, 
Lenient-DQN conceived by \citet{palmer2017lenient} utilizes decaying temperature values for adjusting the policy updates sampled from the experience replay memory. 

\textit{Scalability}: One way to deal with the non-stationarity problem is to train the agents in a centralized fashion and let them act according to a joint policy. However, this approach is not scalable as the number of agents increases, the  state-action spaces grow exponentially, a phenomenon known as "combinatorial complexity" \cite{Kartal2015,hernandez2019survey,yang2020overview}. 
To balance the challenges imposed by non-stationarity and scalability, a centralized training and decentralized execution approach has been proposed \cite{sunehag2017valuedecomposition,rashid2018qmix,lowe2017multi,foerster2017counterfactual,chen2019new}. 

\noindent\textbf{Modeling Multi-agent Reinforcement Problems.}
This section summarizes the common approaches of modeling and solving MADRL problems.

\textit{Independent-learning}: Under this approach each agent considers other agents as part of the environment; consequently each agent is trained independently~\cite{tan1993multi,lauer2000algorithm,tampuu2017multi-agent}. This approach does not suffer from the scalability problem \cite{tampuu2017multi-agent,foerster2017stabilising}, but it makes the environment non-stationary from each agent's perspective~\cite{oroojlooyjadid2019review}. Furthermore, it conflicts with the usage of experience replay that improves the DQN algorithm~\cite{Mnih2015}. To stabilize the experience replay buffer in MADRL settings, \citet{foerster2017stabilising} used importance sampling and replay buffer samples aging.

\textit{Fully observable critic}: 
A way to deal with the non-stationarity of a MADRL environment is by leveraging an actor-critic approach. \citet{lowe2017multi} proposed a multi-agent deep deterministic policy gradient (MADDPG) algorithm, where the actor policy accesses only the local observations whereas the critic has access to the actions, observations, and target policies of all agents during training. As the critic has global observability, the environment becomes stationary even though the policies of other agents change. A number of extensions to MADDPG has been proposed~\cite{chu2017parameter,ryu2018multi,mao2018modelling,wang2020r}.

\textit{Value function decomposition}: Learning the optimal action-value function in fully cooperative MADRL settings is challenging. To coordinate the agents' actions, learning a centralized action-value function, $ Q_{tot} $, is desirable. However, when the number of agents is large, learning such a function is challenging. Independent-learning (where each agent learns its action-value function, $Q_i$) does not face such a challenge, but it also neglects interactions between agents, which results in sub-optimal collective performance. Value function decomposition methods try to capitalize on the advantages of these two approaches. It represents $Q_{tot}$ as a mixing of $Q_i$ that is conditioned only on local information. Value-Decomposition Network (VDN) algorithm assumes that $ Q_{tot}$ can be additively decomposed into  $N Q_i$ for $N$ agents. QMIX~\cite{rashid2018qmix} algorithm improves on VDN by relaxing some of the additivity constrains and enforcing positive weights on the mixer network.

\textit{Learning to communicate}: Cooperative environments may allow agents to communicate. In such settings, the agents can learn a communication protocol to achieve their shared objective more optimally~\cite{kasai2008learning,giles2002learning}. \citet{foerster2016learning} proposed two algorithms, Reinforced Inter-Agent Learning (RIAL) and Differentiable Inter-Agent Learning (DIAL), that use deep networks to learn to communicate. RIAL is based on Deep Recurrent Q-Network with independent Q-learning. It shares the parameters of a single neural network between the agents. In contrast, DIAL passes gradients directly via the communication channel during learning. While a discrete communication channel is used in realizing RIAL and DIAL, CommNet~\cite{sukhbaatar2016learning} utilizes a continuous vector channel. Over this channel, agents obtain the summed transmissions of other agents.
Results show that agents can learn to communicate and improve their performance over non-communicating agents.

\textit{Partial observability}: \citet{foerster2016learning} introduced a deep distributed recurrent Q-network (DDRQN) algorithm based on a long short-term memory network to deal with POMDP problems in the multi-agent setting. \citet{gupta2017cooperative} extended three types of single-agent RL algorithms based on policy gradient, temporal-difference error, and actor-critic methods to the multi-agent systems domain. Their work shows the importance of using DRL with curriculum learning to address the problem of learning cooperative policies in partially observable complex environments. 

We refer the interested reader to the following survey papers for a more in-depth discussion on the topic of multi-agent reinforcement learning: \citet{hernandez2017survey} provide a comprehensive survey on the non-stationarity problem in MADRL; \citet{oroojlooyjadid2019review} scope their survey to include the papers that study decentralized MADRL models with a cooperative goal; \citet{da2019survey} focus on transfer learning for MADRL systems; 
a survey on MADRL from the perspective of challenges and applications is introduced by \citet{du2020survey}; a selective overview of theories and algorithms is presented in~\cite{zhang2019multi-agent}; and a survey and critique of MADRL is given in~\cite{hernandez2019survey}.

\subsubsection{Multi-agent Evolution Strategies}

ES algorithms do not require the problem to be formulated as an MDP; therefore, they do not suffer from the non-stationarity of the environment. Consequently, it is relatively easy to extend a single-agent ES algorithm to the multi-agent domain and develop an application. 
\citet{Hiraga2018} developed robotics controllers based on ESs for managing congestion in robotic swarms path formation using LEDs. The performed experiment covered a swarm of robots, each having seven distance sensors, a ground sensor, an omnidirectional camera, and
RGB LEDs. An artificial neural network (three-layered neural network) represents the controller of the robot, having as inputs: the distance sensors, ground sensors, and the cameras, and as outputs: the motors and LEDs controls. ($\mu,\lambda$)-ES is utilized to optimize the weights of the controller. A copy of the controller is implemented on N different robots, before being evaluated and assessed depending on the swarm's performance. Another similar approach was proposed in \cite{Hiraga2019_swarm} for building a swarm capable of cooperatively transporting food to a nest and collectively distinguishing between foods and poisons. \citet{Hiraga2019_swarm} developed a controller for a robotic swarm using CMA-ES, aiming to automatically generate the behavior of the robots. 

\citet{tang2020learning} proposed an adversarial training multi-agent learning system, in which a quadruped robot (protagonist) is trained to become more agile by hunting an ensemble of robots that are escaping (adversaries) following different strategies. An ensemble of adversaries is used, as each will propose a different escape strategy, thus improving agility (agility refers to coordinated control of legs, balance control, etc.). Training is done using ESs and more specifically by augmenting CMA-ES to the multi-agent framework. There are two steps for training: An outer loop which iteratively trains the protagonist and adversaries, and an inner loop for optimizing the policy of each. Policies are represented by feed-forward neural networks and are optimized with CMA-ES. 

\citet{chen2020framework} proposed a predator-prey system that leverages ESs (OpenAI-ES, CMA-ES). It consists of having multiple predators trained to catch prey in a certain time frame. The predator controllers are homogeneous and are represented by neural networks which parameters are optimized with ESs (OpenAI-ES, CMA-ES) and Bayesian Optimization. The NN has three inputs (the inverse of the distance from the predator to the other nearest predator, the angle between the orientation of the predator and the direction of the prey relative to the predator, the distance between the predator itself and the prey), one hidden layer and two outputs for controlling the angular velocities of the two wheels. As for the prey's controller, it follows a simple fixed evasion strategy: having computed a danger zone map, the prey navigates towards the least dangerous locations. After performing various experiments, the predators showcased a successful collective behavior: moving following a formation and avoiding collisions. 

\textit{Multi-agent credit assignment problem.}
In a multi-agent setting, agents often receive a shared reward for all the agents, making it harder to learn proper cooperative behaviors. \citet{ Li2020Parallel} thus proposed to use Parallelized ESs along with a Value Decomposition Network (useful for identifying each agent's contribution to the training process) for solving cooperative multi-agent tasks. Figure \ref{fig_PES} is an overview of the overall PES-VD algorithm, which consists of two phases. First, the policies of each agent are represented by a NN with parameters $\pmb\theta$, optimized using Parallelized ES. Each agent thus identifies its actions independently following its policy and by interacting with its environment. In a second place, seeing how the reward is common to the whole team, a Value Decomposition Network is used to compute the fitness for each of the different policies. PES-VD is implemented in parallel on multiple cores: $M$ workers evaluate the policies and compute the gradients of the Value Decomposition Network and a master node collects the data and updates the policies and the Value Decomposition Network accordingly. 

\begin{figure}[t]
    \centerline{\includegraphics[width=\columnwidth]{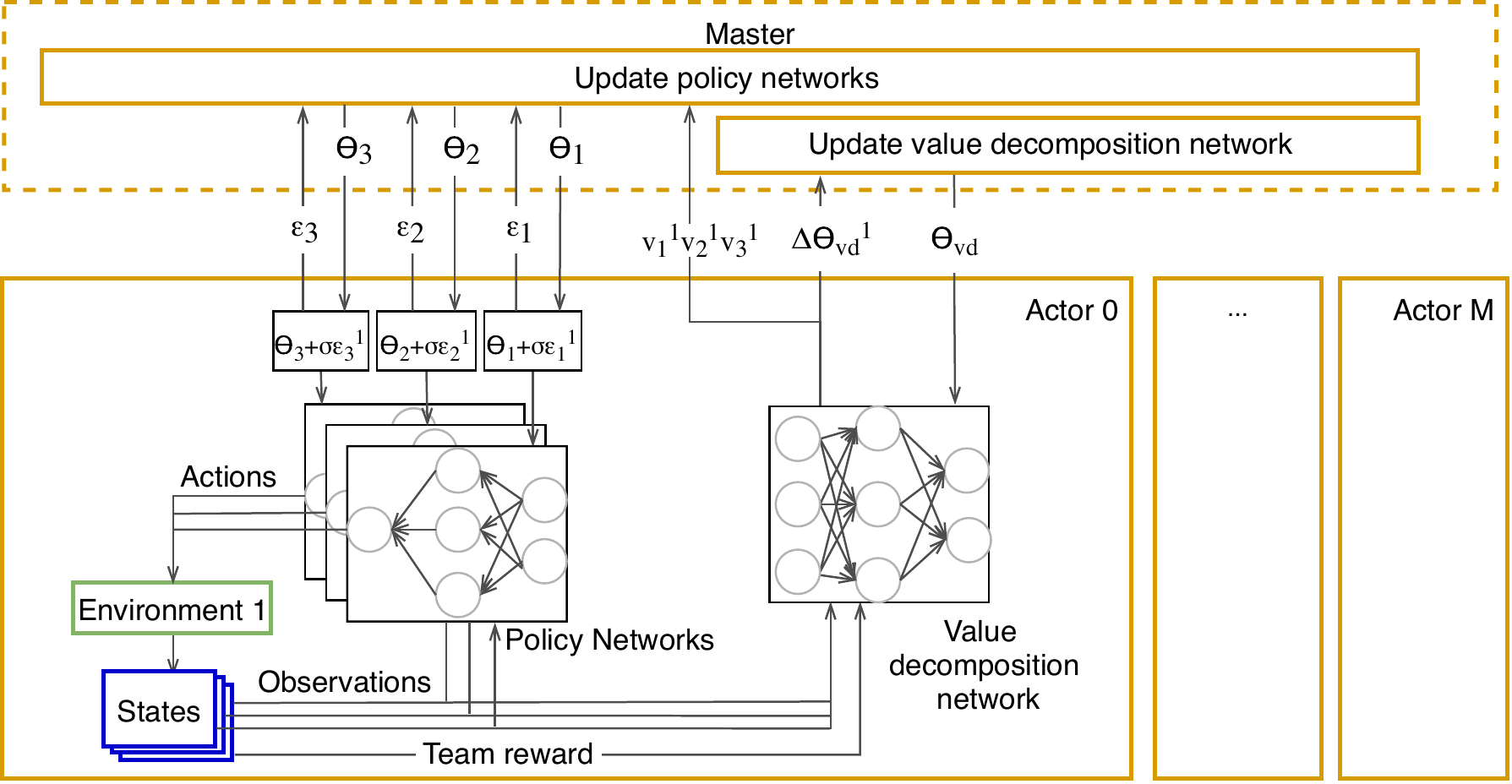}}
    \caption{PES-VD overview} 
    \label{fig_PES}
    \vspacecontrol
\end{figure}

Various researchers proposed multi-agent solutions for swarm scenarios leveraging ESs. Each robot in the swarm runs the same network, thus maintaining collective behavior. \citet{Mart2020} assess the performance of ESs (CMA-ES, PEPG, SES, GA, and OpenAI-ES) for multi-agent learning in the swarm aggregation task. In this problem, the robots controllers are represented by a NN with 2 hidden layers. Each has 8 infrared sensors and 4 microphones for inputs and 2 wheels and a speaker as output. 
Similarly, \citet{fan2018modelbased} used ESs on different multi-agent UAV swarm combat scenarios. \citet{app11062856} developed a swarm foraging behavior using DRL and CMA-ES.

\subsubsection{Comparison}
Here we summarize our observations of this section
\begin{itemize}
    \item Training under a multi-agent setting is more challenging than training a single agent for a plethora of reasons. There are usually two types of agents in MADRL: cooperative and competitive agents. Algorithms can make use of a centralized or decentralized framework and will act in a partially or fully observable environment.
    \item New algorithms such as PES-VD \cite{ Li2020Parallel} propose a direct solution to some of the main challenges of MADRL. PES-VD uses a Value Decomposition Network for solving multi-agent credit assignment problems.
    \item Using ESs for multi-agent learning is still a growing field with a large potential as to the many advantages ESs can bring to concepts such as ``collective robotic learning" and ``cloud robotics" \cite{gu2016deep} with its improved approach to parallelism \cite{salimans2017evolution}. 
    \item Semi-centralized MADRL systems are an interesting future research direction in which a few central entities possess partial information of a set of agents. 
    \item The literature on ESs for multi-agent scenarios seems to focus on enabling applications. We hypothesis that this is because it is less challenging to extend single-agent ES algorithms to the multi-agent domain. This is because ES algorithms do not require Markov property in the formulation of the problem, and therefore, they do not suffer from non-stationary environments.   
\end{itemize}

\section{Hybrid Deep Reinforcement Learning and Evolution Strategies Algorithms}
\label{sec:hybrid}
\begin{figure}[t]
    \centering
    \includegraphics[width=\columnwidth]{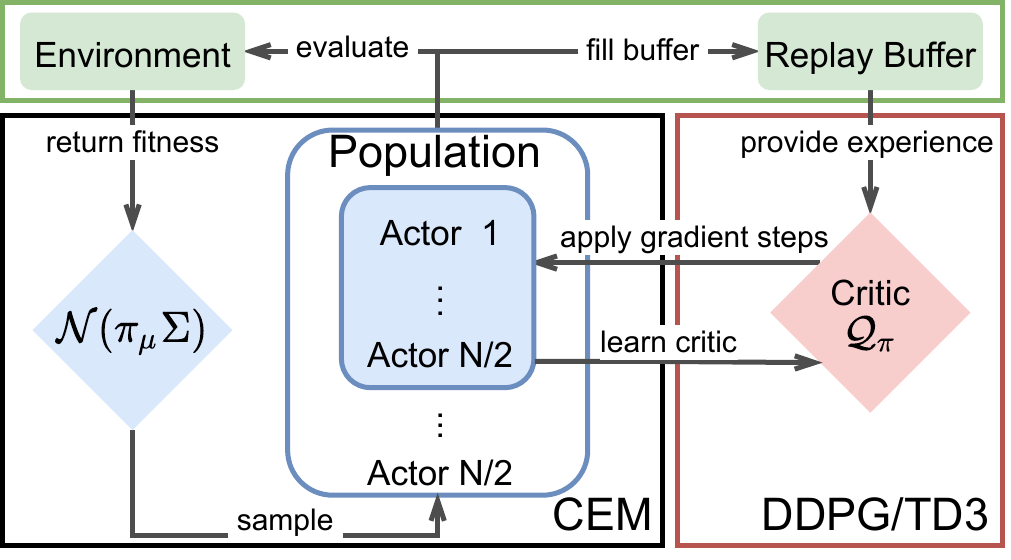}
    \caption{CEM-RL~\cite{pourchot2019cemrl}: a hybrid algorithm that combines cross-entropy method with (Twin) Deep Deterministic Gradient Policy~\cite{fujimoto2018addressing}.}
    \label{fig:cem-rl}
    \vspacecontrol
\end{figure}

Although DRL and ES have the same objective---optimizing an objective function in a potentially unknown environment---they have different strengths and weaknesses \cite{Hansel2021,ecoffet2021policy}. 
For example, DRL can be sample efficient thanks to the combination of RL and deep learning; while ES have robust convergence properties and exploration strategies.
The hybrid approach combines DRL and ES to get the best of both worlds. 
Although the idea is not new~\cite{stafylopatis1998autonomous}, hybridizing DRL and ES has gained momentum, driven by the recent success of DRL and ES~\cite{salimans2017evolution,jaderberg2019human}. Combining these strengths has, among others, let to very strong play in the most challenging Real-Time Strategy (RTS) games such as StarCraft \cite{alphastarblog}. We describe in the following a few population-guided parallel learning schemes that enhance the performance of RL algorithms.

\citet{pourchot2019cemrl} addressed the problem of policy search by proposing CEM-RL: a hybrid algorithm that combines a cross-entropy method (CEM) with either the Twin Delayed Deep Deterministic policy gradient (TD3)~\cite{fujimoto2018addressing} or the Deep Deterministic Policy Gradient DDPG~\cite{lillicrap2019continuous} algorithms (Figure~\ref{fig:cem-rl}). The CEM-RL architecture consists of a population of actors that are generated using CEM, and of a single DDPG or TD3 agent. The actors generate diversified training data for the DDPG/TD3 agent, and the gradients obtained from DDPG/TD3 are periodically inserted into the population of the CEM to optimize the searching process. 
The authors showed that CEM-RL is superior to CEM, TD3~\cite{fujimoto2018addressing}, and Evolution Reinforcement Learning (ERL)~\cite{khadka2018evolutionary}: a hybrid algorithm that combines a DDPG agent with an evolutionary algorithm.
\citet{shopov2018} combined ESs and multi-agent DRL (Deep Q-Networks) for Sequential Games and showcased the model's efficiency as compared to Classical multi-agent reinforcement training with $\epsilon$-greedy. The experiment performed by \citet{shopov2018} aims to optimize the behaviour of a group of autonomous agents (the pursuers) in a map. Tests were performed on two cases: one map with almost no obstacles and another with many obstacles (increased probability of falling into the local minimum). Using ESs on the latter yielded better performance.

\citet{houthooft2018evolved} devised a hybrid RL agent, Evolved Policy Gradients (EPG), that, in addition to the policy, optimizes a loss function. EPG consists of two optimization loops: the inner loop uses stochastic gradient descent to optimize the agent's policy, while the outer one utilizes ES to tune the parameters of a loss function that the inner loop minimizes. Thanks to this ability to fine tune the loss function according to the environment and agent history, EPG can learn faster than a standard RL agent. 

\citet{Chen2020} proposed a hybrid agent to approximate the Pareto frontier uniformly in a multi-objective decision-making problem. The authors argued that despite the fast convergence of DRL, it cannot guarantee a uniformly approximated Pareto frontier. On the other hand, ES achieve a well-distributed Pareto frontier, but they face difficulties optimizing a DNN. Therefore, \citet{Chen2020} proposed a two-stage multi-objective reinforcement learning (MORL) framework. In the first stage, a multi-policy soft actor-critic algorithm learns multiple policies collaboratively. And, in the second stage, a multi-objective covariance matrix adaptation evolution strategy (MO-CMA-ES) fine-tunes policy-independent parameters to approach a uniform Pareto frontier. 

\citet{Bruin2020IFAC} used a hybrid approach to train and fine-tune a DNN control policy. Their approach consists of two main steps: (i) learning a state representation and initial policy from high-dimensional input data using gradient-based methods (i.e., DQN or DDPG); and (ii) fine-tuning the final action selection parameters of the DNN using CMA-ES. This architecture enables the policy to surpass in performance its gradient-based counterpart while using fewer trials compared to a pure gradient-free policy.

Several other researchers have also proposed solutions hybridizing ES and DRL for various applications. For example,  \citet{song2021esenas} proposed ES-ENAS, a neural architecture search (NAS) algorithm for identifying RL policies using ES and Efficient NAS (ENAS);  \citet{ferreira2021learning} used ES to learn agent-agnostic synthetic environments (SEs) for Reinforcement Learning. 
\subsection{Comparison}

\begin{table*}[t]
\centering
    \caption{Hybrid algorithms highlights}
    \begin{tabular}{p{2cm}|p{6.6cm}|p{7.4cm}|p{0.4cm}}
         \hline 
          \textbf{Algorithm} & \textbf{Description} & \textbf{Experiments} & \textbf{Ref.}\\
          \hline
                    CEM-RL &  combines a cross-entropy method and Twin Delayed Deep Deterministic policy gradient~\cite{fujimoto2018addressing} to find robust policies &  
                    outperforms CEM, TD3, multi-actor TD3, and Evolutionary Reinforcement Learning~\cite{khadka2018evolutionary} & \cite{pourchot2019cemrl} \\
          \hline
                    Evolved Policy Gradients (EPG) & uses gradient descent and CMA-ES for policy and loss function optimization, respectively & achieves faster learning than policy gradient methods and provides qualitatively different behavior from other popular meta-learning algorithms & \cite{houthooft2018evolved}\\
          \hline
                    MO-CMA-ES & integrates a multi-policy soft actor-critic algorithm with a multi-objective covariance matrix adaptation evolution strategy to approach uniform Pareto frontier& 
                    exceeds other algorithms such as the hypervolume-based~\cite{Moffaert2013}, radial~\cite{Parisi2014}, Pareto following~\cite{Parisi2014}, and Deep Neuroevolution~\cite{such2018deep} algorithm on computing the Pareto frontier & \cite{Chen2020}\\
          \hline
                  Fine-tuned DRL & combines CMA-ES and DQN or DDPG to train and fine-tune a DNN control policy & 
                  surpasses gradient-based methods while requiring less iterations than gradient-free ones & \cite{Bruin2020IFAC}\\ 
          \hline
    \end{tabular}
    \label{tab:hyb}
\end{table*}
Here we summarize our observations of this section:
\begin{itemize}
    \item DRL suffers from temporal credit assignment, sensitivity in the hyperparameters' selection and might suffer from more brittle exploration due to its unique agent setting, while ES has low data efficiency and struggle with large optimization tasks.
    \item Combining both approaches can help address some of these identified challenges.
    \item Some hybrid methods proposed throughout the literature seem to outperform the use of each method on its own.
    \item Hybridizing DRL and ES is still a relatively new field of research. 
\end{itemize}

\section{Applications}
\label{sec:applications}
Next, we compare DRL and ESs based on the applications they support. The goal is to get an indication of their potential by tracking their application record so far. The results of querying Google Scholar is presented in Table~\ref{table:search_terms} in conjunction with the keywords that were used\footnote{The search query template is ``allintitle: ``evolution strategies'' OR ``evolutionary strategies'' \text{key\_word\_1 OR key\_word\_1 -excluded\_key\_word}'' and ``allintitle: ``reinforcement learning'' \dots}.

\subsection{Deep Reinforcement Learning applications}

%

%
\begin{table}[t]  
\footnotesize
\setlength{\tabcolsep}{1pt} 
\setlength\extrarowheight{1pt}
\caption{Searching Google Scholar for DRL and ESs applications (only papers' titles are considered).  }
\label{table:search_terms} 
\begin{tabularx}{\columnwidth}{|p{2cm}|X|p{.6cm}|p{.6cm}|}
    \hline  
    \multirow{2}{*}{\textbf{Industry field}} & \multirow{2}{*}{\textbf{Search terms}} & \multicolumn{2}{c|}{\textbf{Results}} \\
    \cline{3-4}
     & & \textbf{DRL} & \textbf{ESs} \\
    \hline
    \textit{Gaming} & game, games, gaming, playing, play, mahjong, atari, tetris, soccer \textit{excluding} survey and review  & 1630 &  44\\
    \hline
    \textit{Robotics} & robotics, ``motion control'', robots, ``robot navigation'', assembly, robot, grasping \textit{excluding} survey and review & 2350 & 39 \\
    \hline
    \textit{Finance} & finance, financial, trading, portfolio, stock, price, liquidation, hedging, banking, trader, cryptocurrency, underpricing \textit{excluding} survey and review  & 475 & 34 \\
    \hline
    \textit{Communications} & network, routing, communications, wireless, 5g, LTE, MAC, ``access control'', ``network slicing'', \textit{excluding} ``neural network'' survey and review & 2020 & 53\\
    \hline
    \textit{Energy} & energy, power \textit{excluding} survey and review & 1470 & 41 \\
    \hline
    \textit{Transportation} & transportation, transport, vehicle, traffic, fleet, driving \textit{excluding} survey and review & 1580 & 25 \\
    \hline
\end{tabularx}
\end{table}
 
\noindent\textbf{Gaming.} Video games such as the Atari games \cite{bellemare2013arcade} are excellent testbeds for DRL algorithms, given their well-defined problem settings and virtual environment. This makes evaluation safe and fast compared to real-world experiments \cite{shao2019survey}.

There have been two important triumphs for DRL with respect to perfect information games. First, in 2015, \citet{Atari2015} developed an algorithm that could learn to play different Atari 2600 games at a superhuman level using only the image pixels as input. This work paved the way for DRL applications trained on high-dimensional data based only on the reward signal. Soon after, in 2016, \citet{alphaGo} developed AlphaGo, the first program ever to beat a world champion in Go. Instead of the handcrafted rules often seen in chess programs, AlphaGo consisted of neural networks trained using a combination of Supervised Learning (SL) and RL. Only a year later, this achievement was triumphed by \citet{silver2017mastering}, whose AlphaGo Zero program beat its predecessor AlphaGo. AlphaGo Zero was based solely on RL, omitting the need for human data. More recent works have also been successful in imperfect information games which, unlike Go and Atari games, only let agents observe part of the system. In OpenAI Five~\cite{openai2019dota},  agents were able to defeat the world's esports champions in the game of Dota2, while AlphaStar \cite{vinyalsStarcraft} attained one of the highest rankings in the complex real-time strategy game of StarCraft II. \cite{nair2015massively,espeholt2018impala,badia2020up,bellemare2016unifying} further examined DRL algorithms' ability to scale, parallelize, and explore using Atari games. Lastly, an extensive survey on  DRL in video games has been composed by \citet{shao2019survey}.

\noindent\textbf{Robotics} is another domain which forms a prominent testbed for DRL algorithms \cite{Kober2013,Nguyen2019}. DRL can provide robots with navigation, obstacle avoidance and decision making capabilities, by mapping sensory data directly to actual motor commands~\cite{chen2017decentralized,kahn2018self}. In some cases this has enabled robots to learn complex movements such as jumping or walking ~\cite{bellegarda2020robust, Haarnoja2018}. \citet{tai2017robots} proposed a mapless motion planner which relies on training in simulation, after which, physical agents were able to navigate unknown static environments without fine-tuning. While most works involved simulation, \citet{gu2016deep} showed that DRL can be used to learn complex robotics 3D manipulation skills from scratch on real-world robots and further reduced training time by parallelizing the training across multiple robots. \citet{Haarnoja2018} demonstrated that using DRL, one can also achieve stable quadrupedal locomotion on a physical robot within a reasonable time without prior training. 
For an in-depth review of the use of DRL for robot manipulation, we refer the interested reader to \cite{Kober2013,Nguyen2019}.

\noindent\textbf{Finance.} DRL also finds applications in trading \cite{Li2019trading, Deng2017trading} and investment management \cite{jiang2017portofolio}, including cryptocurrency \cite{jiang2017crypto}.
\citet{Moody1998ReinforcementLF} built a DRL agent for stock trading using raw financial data as the DNN input. \citet{CARAPUCO2018783} described a system for short-term speculation in the foreign exchange market, based on DRL. \citet{WU2020142} proposed adaptive stock trading strategies leveraging DRL. A more recent DRL work by \citet{Lei2019finance}, adaptively selects between historical data and the changing trend of a stock, depending on the current state.


\noindent\textbf{Communications.} Upcoming networks such as the 5G network, emphasize the need for efficient dynamic and large-scale solutions \cite{XiongNetCommunications}. 
DRL has been emerging as an effective tool to tackle various problems and challenges within the field of networking \cite{LuongHoangNetSurvey}. For example, \citet{WangNet2018} applied a DQN to automatically optimize data transmission and reception in a multi-wireless-channel access problem. \citet{YeNet2017} developed a similar system for vehicle-to-vehicle communication. The optimal transmission bitrate can change over time. DRL 
can dynamically optimize the bitrate based on the quality of the last segment, the current buffer state \cite{GadaletaChiariottiNet,HongziRaviNet}
and other channel statistics \cite{Chinchali2018CellularNT,FerreiraPaulo}. 
Proactive caching can greatly reduce the number of transmissions over the network.
However, deciding which content to cache is not trivial. 
Researchers have used DQNs to determine which information to keep in a cache based on observations of the channel state \cite{HeHuNet}, cache state \cite{HeLiangYuNet}, request history \cite{ZhongGursoyNet,HeZhangYuNet} and available base stations \cite{YingYuZhaoNet,HeLiangzZhangNet,YingZhaoYinNet}. 



\noindent\textbf{Energy.} Within the energy sector, \textit{smart grids} make intelligent decisions with respect to electricity generation, transmission, distribution, consumption and control. DRL has been used in a variety of settings to tackle electric power system decision and control problems \cite{glavic2017reinforcement}, such as in the context of microgrids \cite{franccois2016deep} or buildings energy optimization \cite{mocanu2018line,ruelens2019direct}.
 
\noindent\textbf{Transportation.} Congestion, safety and efficiency are important aspects of transportation. DRL is often used for adaptive traffic signal control to reduce waiting times \cite{li2016traffic,liang2019traffic,Chu2020traffic}. \citet{Chen2020traffic} expanded upon this and conceived the first DRL control system which scales to thousands of traffic lights. \citet{Wang2020bus} developed a MADRL framework to prevent `bus bunching' and streamline the flow of public transport. \citet{Manchella2021transport} proposed a model-free DRL algorithm which packs ride-sharing passengers together with goods delivery to optimize fleet utilization and fuel efficiency. 

\subsection{Evolution Strategy applications}
ESs applications are categorized and highlighted next. 

\noindent\textbf{Gaming.} similarly to DRL, gaming represents one of the main testbeds for ESs. Most of the literature on ESs reviewed in this survey test their algorithms on Atari games \cite{salimans2017evolution, pagliuca2020efficacy,chrabaszcz2018basics,zhou2019sample,chen2019restart,conti2018improving, risi2019deep,ijcai2019}. These are considered to be challenging as they present the agents with high dimensional visual inputs and a diverse and interesting set of tasks that were designed to be difficult for humans players \cite{mnih2013playing}.

\noindent\textbf{Robotics.} The ability of ESs to continuously control actuators has been leverage in controlling simulated and real robotic systems \cite{ salimans2017evolution,Liu2019TrustRE,zhang2020accelerating,Liu_2020HistoricalES,veer2020cones,Shi2020maxEntropy,Jackson2019}. \citet{hu2018evolution} used CMA-ES to make a robot learns how to grasp object under uncertainty. \citet{uchitane2010evolution} augmented the ($\mu+\lambda$)-ES algorithm with a mask operation during the mutation step to tune the controller's parameters of humanoid robots. With help of ESs \citet{li2008novel} design an indoor mobile robot navigation using monocular vision.  


\noindent\textbf{Finance.} \citet{Korczak2001} presented a portfolio optimization algorithm using ESs. \citet{Rimcharoen2005,Sutheebanjard2009} proposed the Adaptive and (1+1)-ES methods for predicting the Stock Exchange of Thailand index movement. \citet{Bonde2012StockPP} predicted the changes (increase or decrease) of stock prices for different companies using ESs and Genetic Algorithms. \citet{Vijayalakshmi2012} proposed ESs with hall of fame (ES-HOF) for optimizing long–short portfolios with the 130-30-strategy-based constraint. \citet{Vijayalakshmi2014} used multi-objective ESs for futures portfolio optimization. \citet{Yu2017DynamicPO} proposed an ESs method  for the multi-asset multi-period portfolio optimization. \citet{Sable2017} proposed an ESs approach for predicting the short time prices of stocks. \citet{Sorensen2020} applied meta-learning algorithms to ES for stock trading.

\noindent\textbf{Communications.} Different methods are proposed throughout the literature that used ESs for communication. \citet{Perez2007} used ESs with NSGAII (ESN) to approximate the Pareto frontier of the mobile adhoc network (MANETs). \citet{Krulikovska2010} used ESs for the routing of multipoint connections. Additionally, they proposed methods for improving ESs. \citet{Nissen2008} used ESs for designing a survivable network while taking economics and reliability into consideration. \citet{He2018} analyzed the data characteristics of wireless sensor network (WSN), and proposed a method for fault diagnosis of WSN based on a belief rule base (BRB) model which is optimized using CMA-ES.\citet{SrivastavaS18} used ESs for solving the the total rotation minimization problem (TRMP) in directional sensor networks. \citet{Srivastava2020AnES} presented an ESs method for solving the Cover scheduling problem in wireless sensor networks (WSN-CSP). \citet{Gu2021} proposed an ESs method to search for a network configuration able to produce and stabilize responses of a Physical Unclonable Functions (PUFs).

\noindent\textbf{Energy.}
ESs have been used to optimize many energy-related systems. For example, \citet{mendoza2006optimal} used ESs to select optimal size of feeders in radial power distribution system. \citet{lezama2017differential} used differential ESs for large-scale energy resource management in smart grids. \citet{coelho2014heuristic} used it for energy load forecasting in electric grids.  \citet{versloot2014optimization} optimized near-field wireless power transfer using ESs. 

\noindent\textbf{Transportation.} A number of papers that use ESs for managing and optimizing  vehicle traffic have been proposed.
\citet{Balaji2007} proposed a multi-agent-based real-time centralized evolutionary optimization technique for urban traffic management in the area of traffic signal control. \citet{mester2005active} combined ESs with guided local search to tackle large-scale vehicle routing problems with time windows. \citet{mester2005active} simulated and optimized traffic flow with help of ESs. 

\subsection{Comparison}
Next we list our observations about this section.
\begin{itemize}
    \item Both DRL and ES algorithms have found adoption in many domains such as robotics, games, and finance.
    \item DRL-based solutions seem to excel in situations that require scalable and adaptive behavior. 
    \item DRL receives much more attention than ESs within the scientific (Table~\ref{table:search_terms}). We suspect that two of the main reasons for this gap are (i) DRL has a richer structure; therefore, it naturally allows for more research, and (ii) there are similar algorithmic families to ESs (e.g., genetic algorithms) which may result in reduced focus on ESs.
    \item Many great-performing DRL and ES algorithms are benchmarked in simulated environments. Consequently, their performance in real-world applications are still questionable. 
\end{itemize}

\section{Challenges and future research directions}
\label{sec:future_direction}
Although DRL and ESs have proven their worth in many AI fields, there are still many challenges to be addressed. We briefly list some of them in the sequel. 

\noindent \textbf{Sample Efficiency.} 
DRL agents require a large number of samples (i.e., interactions with environments) to learn good-performing policies. Collecting so many samples is not always feasible due to either computational reasons or because the quantity of interactions with the environment is limited. Although this problem has been tackled in different ways (e.g., transfer learning, meta-learning), more innovation and research are still needed~\cite{nagabandi2020deep,tebbe2020sample}. One promising research direction to tackle this problem is model-based RL. However, getting an accurate model of the environment is usually hard. 

ESs can provide more robust policies as compared to DRL; however, they are even less sample efficient, as they work with full-length episodes \cite{pourchot2018importance,sigaud2019policy}, and they do not use any type of memory \cite{pourchot2018importance}. Approaches to improve sample efficiency in ESs such sample resue and importance mixing \cite{sun2012efficient,pourchot2018importance} have been proposed; however, more research and innovation are still required. 

\noindent\textbf{Exploration versus exploitation.} The exploration versus exploitation dilemma is one of the most prominent problems in RL. Beyond classical balancing approaches such as $\varepsilon$-greedy~\cite{sutton2018reinforcement}, Upper Confidence Bound (UCB)~\cite{auer2002finite}, and Thompson Sampling~\cite{russo2017tutorial}, recent breakthroughs enable the exploration of novel environments. For example, \citet{osband2016deep} observed the importance of temporal correlation and proposed the bootstrapped DQN; and \citet{bellemare2016unifying} used density models to scale UCB to problems with high-dimensional input data. In spite of that, exploring complex environments is still a very active field of research.  

ESs realize exploration through the  recombination and mutation steps.  
Despite their effectiveness in exploration, ESs may still get trapped in local optima~\cite{Liu2019TrustRE, zhang2020accelerating}. Proposals have been made to enhance ESs exploration capabilities \cite{choromanski2018structured,maheswaranathan2019guided}; however, more work in this direction is needed. In general, DRL and ESs are proposed to tackle ever more novel environments, and consequently, the exploration versus exploitation dilemma still poses a challenge that requires innovation. 

\noindent \textbf{Sparse reward.} 
A reward signal guides the learning process of an RL agent. When this signal is sparse learning becomes much harder.
Although, different solutions have been introduced (e.g., reward shaping~\cite{sutton2018reinforcement}, curiosity-driven methods~\cite{pathak2017curiositydriven}, curriculum learning~\cite{portelas2020automatic}, hierarchical learning~\cite{Arulkumaran_2017} and inverse RL \cite{Andrew2000}), learning with sparse rewards still represents an open challenge.  

Direct the exploration of ES algorithms to counter the sparsity and/or deceptiveness of an RL task is one of the most important challenges to scale ESs to more complex environments and make them more efficient. 
A detailed summary of the challenges related to ESs, such as differential evolution and swarm optimization, is presented in \cite{Zhenhua2020}.

\noindent\textbf{Simulation-to-reality gap.} Despite the benefits of simulations, they give rise to the sim-to-real gap: policies that are learned in simulations often do not work as expected in the real world. Different techniques are being adapted to mitigate the effect of this gap. For example, \cite{sadeghi2016cad2rl,matas2018sim} randomized the simulated environment to produce more generalized models. \citet{rao2020rl} noted that such randomization requires manually specifying which aspects of the simulator to randomize. Therefore, they used domain adaptation (i.e., many simulated examples and a few real ones) to train a robot on grasping tasks without manually instrumenting the simulator. Despite such efforts, the sim-to-real gap is still an open challenge to be addressed.


\section{Conclusion}
\label{sec:conclusion}
Deep Reinforcement Learning (DRL) and Evolution Strategies (ESs) have the same objective but make use of different mechanisms for learning sequential decision-making tasks. In this paper, we provided the necessary background of DRL and ESs in order to understand their relative strengths and weaknesses, which may lead to developing an algorithmic family that is superior to each one of them. Instead of focusing on individual algorithms, we considered major learning aspects such as parallelism, exploration, meta-learning, and multi-agent learning. 
We believe that hybridizing DRL and ESs has a high potential to drive the development of agents that operate reliably and efficiently in the real world.

\section*{Acknowledgment}
This work has been undertaken in the Internet of Swarms project sponsored by Cognizant Technology Solutions and Rijksdienst voor Ondernemend Nederland under PPS O\&I.

\bibliographystyle{IEEEtranN}
\bibliography{main}

\end{document}